\pdfoutput=1

\documentclass[11pt]{article}

\usepackage{emnlp2021}

\usepackage{times}
\usepackage{latexsym}
\usepackage{xspace,mfirstuc,tabulary}
\usepackage[utf8]{inputenc}
\usepackage{times}
\usepackage{latexsym}
\usepackage[ruled,linesnumbered]{algorithm2e}
\usepackage{url}
\usepackage{paralist}
\usepackage{multicol,graphicx,mathrsfs,pifont,amscd,latexsym,color,url, pifont, subfigure,epstopdf,setspace,multirow,colortbl, bbm,listings, balance,color,indentfirst,enumitem, subfigure,amsmath,amssymb, verbatim,microtype}
\usepackage{graphicx,booktabs, paralist}
\usepackage[T1]{fontenc}
\usepackage{caption}
\usepackage[utf8]{inputenc}
\usepackage{amsfonts}
\usepackage{microtype}

\usepackage{amssymb}
\usepackage{pifont}
\newcommand{\cmark}{\ding{51}}%
\newcommand{\xmark}{\ding{55}}%
\usepackage{changepage}

\newcommand{\kg}{Grounded Relational Wiki-Graph\xspace}
\newcommand{\method}{RGPT-QA\xspace}

\usepackage[hybrid]{markdown}

\title{Relation-Guided Pre-Training for Open-Domain Question Answering}

\author{
	Ziniu Hu,
	Yizhou Sun,
	Kai-Wei Chang \\
	University of California, Los Angeles\\
	{\tt \{bull, yzsun, kwchang\}@cs.ucla.edu} \\
}

\begin{document}
\maketitle

\begin{abstract}

Answering complex open-domain questions requires understanding the latent relations between involving entities. However, we found that the existing QA datasets are extremely imbalanced in some types of relations, which hurts the generalization performance over questions with long-tail relations. To remedy this problem, 
in this paper, we propose a Relation-Guided Pre-Training (\method) framework\footnote{Dataset and code are released at \url{https://github.com/acbull/RGPT-QA}.}. We first generate a relational QA dataset covering a wide range of relations from both the Wikidata triplets and Wikipedia hyperlinks. 
We then pre-train a QA model to infer the latent relations from the question, and then conduct extractive QA to get the target answer entity.
We demonstrate that by pre-training with propoed \method techique, the popular open-domain QA model, Dense Passage Retriever (DPR), achieves 2.2\%, 2.4\%, and 6.3\% 
absolute improvement in Exact Match accuracy  on Natural Questions, TriviaQA, and WebQuestions. Particularly, we show that \method improves significantly on questions with long-tail relations.

\end{abstract}

\section{Introduction}\label{sec:introduction}Open domain question answering is a challenging task that answers factoid questions 
based on evidence in a large corpus (e.g., Wikipedia). Most open-domain QA systems follow retriever-reader pipeline~\cite{DBLP:conf/acl/ChenFWB17}, in which a \textit{retriever} selects a subset of candidate entities and associated passages from the corpus that might contain the answer, then a \textit{reader} extracts a text span from the passages as the answer. 
This process involves multiple entities that are relevant to answer the question. The QA system is required to extract these entities from the question and passages and identify the (latent) semantic relations between these entities in order to answer the question. 
For example, to answer the following question: ``Where did Steph Curry play college basketball at?'', the QA model is required to reason the implicit relation triplet $\langle $\textit{Steph Curry, Educated At, Davidson College}$ \rangle$ to identify the correct answer.



To capture the relation knowledge required to answer questions, most QA systems rely on human-annotated supervised QA datasets. However, it is expensive and tedious to annotate a large set of QA pairs that cover enough relational facts for training a strong QA model.
In addition, we showed that even for a large QA dataset like Natural Questions~\cite{DBLP:journals/tacl/KwiatkowskiPRCP19}, its training set only covers 16.4\% of relations in WikiData~\cite{DBLP:journals/cacm/VrandecicK14} knowledge graph. Moreover, for those covered relations, the frequency distribution is imbalanced, i.e., 30\% of relation types appear only once.
Consequently, for the questions involving infrequent (a.k.a, long-tail) relations in the training set, the QA exact match accuracy is 22.4\% lower than average. Such a biased relation distribution of existing QA datasets severely hurts the generalization of trained QA systems.

To improve the open-domain QA systems for questions with long-tail relations,
in this paper, we propose \method, a simple yet effective Relation-Guided Pre-training framework for training QA models with augmented relationa facts from knowledge graph.
The framework consists of two steps: 1) generate a relational QA dataset that covers a wide range of relations without human labeling; 2) pre-train a QA model to predict latent relations from questions and conduct extractive QA.

\begin{figure*}[t!]
   \begin{minipage}{0.48\textwidth}
     \centering
     \includegraphics[width=1.0\columnwidth]{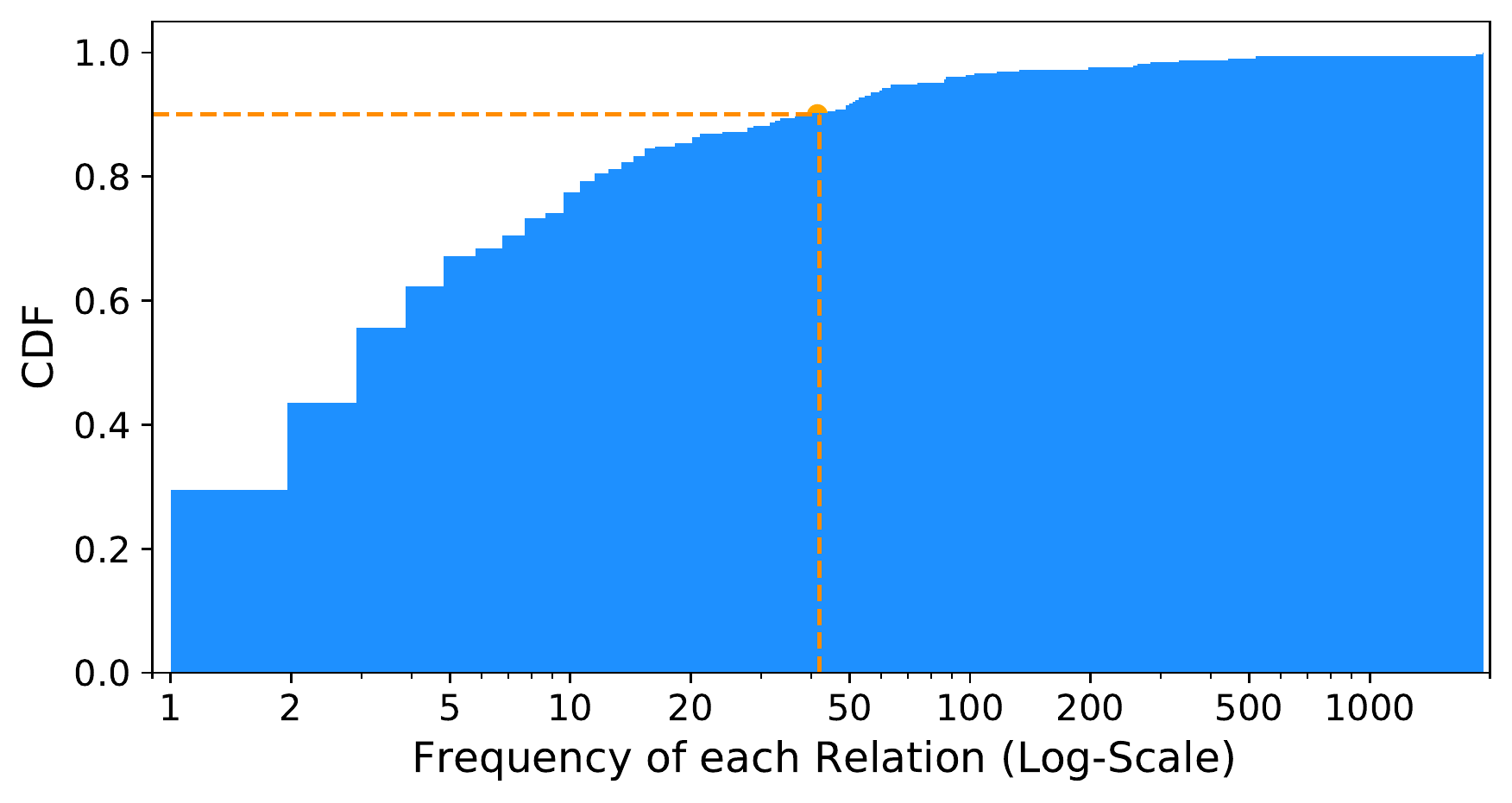}
    \caption{Cumulative distribution function (CDF) of relation frequency in Natural Question Training set.}\label{fig:cdf}
   \end{minipage}\hfill
   \begin{minipage}{0.48\textwidth}
     \centering
     \includegraphics[width=1.0\columnwidth]{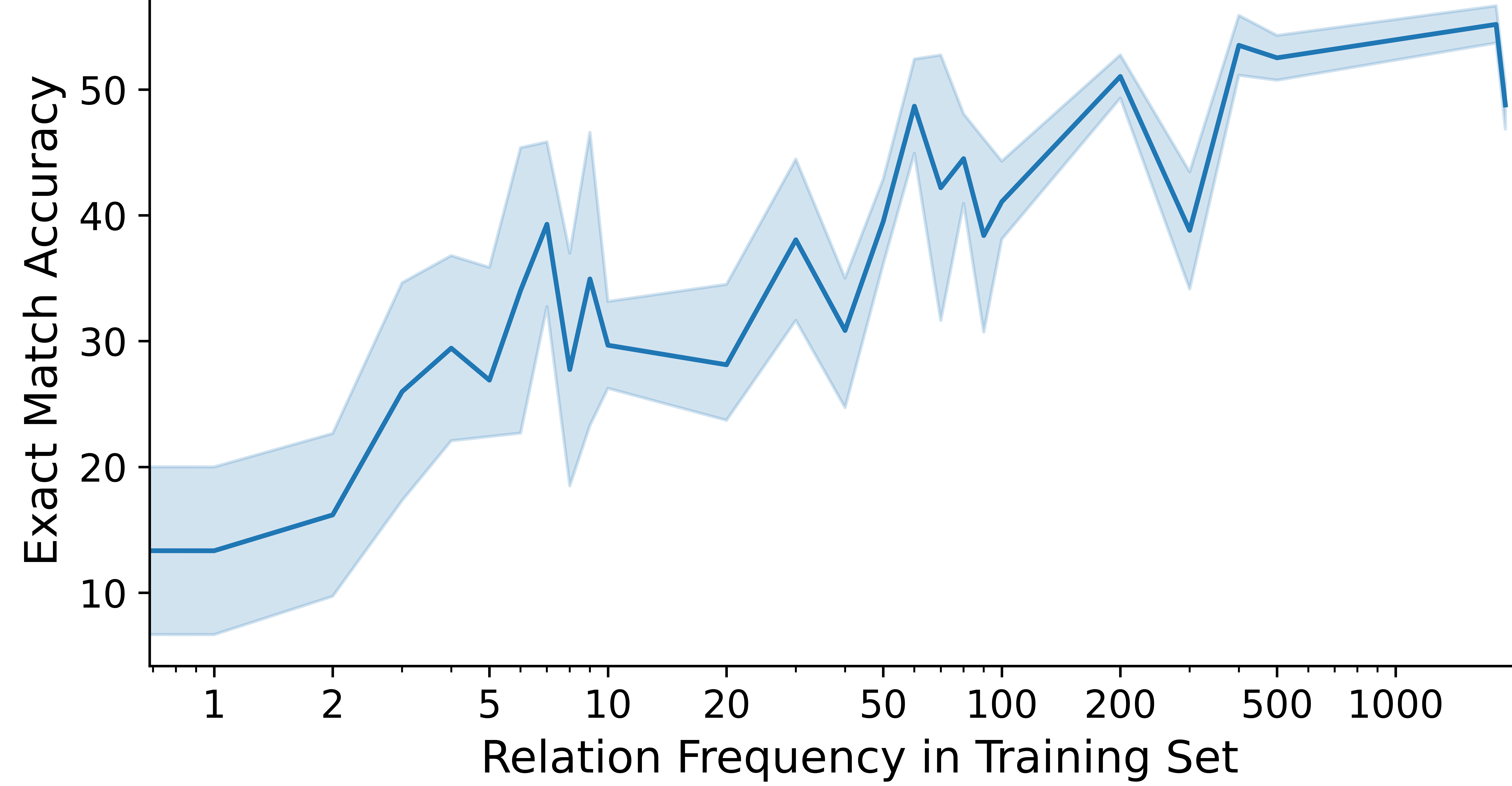}
    \caption{Exact Match accuracy of a trained DPR model in validation set with different relation frequency in training set.}\label{fig:emp}
   \end{minipage}
\end{figure*}


The key of our framework is to generate a relational QA dataset that align entities in Wikipedia passages with structured knowledge graph (e.g., WikiData). We call such a dataset \kg. In this graph, each edge indicates the relationship of two connected entities, and the edge is linked to a passage in Wikipedia describing this relationship. 
As WikiData knowledge graph also suffers from low coverage of long-tail entities and relations,
we further convert hyperlinks in Wikipedia into knowledge triplets without specifying relation labels.
Next, we link each relation triplet to a Wikipedia passage to help generate natural questions. We assume that if one passage in the Wiki-page of source entity contains the target entity, then the context in this passage describes the relationship between the two entities. With the constructed graph, we use a template to synthesize question and answer pairs and then pre-train the QA model to capture the relational facts for answering complex open-domain questions.

As a pre-training method, \method can be incorporated with any open-domain QA system. In this paper, we utilize the recently developed Dense Passage Retriever (DPR)~\cite{DBLP:journals/corr/abs-2004-04906} as the base QA system to evaluate the proposed pre-training effectiveness. Experimental results show that \method enhances DPR's Exact Match accuracy by 2.2$\%$, 2.4$\%$, and 6.3$\%$ on Natural Questions, TriviaQA and WebQuestions respectively. Compared with the existing QA pre-training methods~\cite{DBLP:conf/acl/LeeCT19, DBLP:journals/corr/abs-2002-08909, DBLP:conf/acl/LewisDR19}, \method explicitly captures a wide range of relational facts and thus achieves better performance.
Moreover, for the questions containing long-tail relations in Natural Questions, the performance is improved by 10.9$\%$, showing that \method alleviates the unbalanced relation distribution problem in the existing QA datasets.


The key contributions of this paper are:
\begin{compactitem}
\item We propose \method, a pre-training method to inject knowledge from relational facts in knowledge graph into QA models.
\item \method enhances the performance of a popular QA model, i.e., DPR, especially on the questions with long-tail relations.
\end{compactitem}

\section{Preliminary and Empirical Analysis}\label{sec:preliminary}In this section, we firstly introduce the retriever-reader pipeline for open-domain QA, and then we analyze how the relation distribution in existing QA datsets influence generalization performance.

\paragraph{Open-Domain Question Answering.} We focus on open-domain question answering that requires to extract answer from a large corpus (e.g. Wikipedia) $\mathbb{C}=\{p_i\}_{i=1}^{N}$ containing $N$ passages. 
Most open-domain QA systems follow a retriever-reader pipeline proposed by~\citet{DBLP:conf/acl/ChenFWB17}. Given a factoid question $q$, the QA system first retrieves $K$ relevant passages $\{p_j\}_{j=1}^{K}$ from the corpus $\mathbb{C}$. Then a reading comprehension module extracts a text span $w_{\text{start}}$, \ldots,  $w_{\text{end}}$ from one of these retrieved passages as the answer $a$ to the question. 
Some QA dataset annotated the passage where the answer $a$ is derived. We called this passage ground truth passage.



For the retriever, 
earlier systems utilize term-based retrieval methods, such as TF-IDF and BM25, which fails to capture the semantic relationship between question and passage beyond lexical matching. 
Recent studies~\cite{DBLP:conf/acl/LeeCT19,DBLP:journals/corr/abs-2004-04906, DBLP:conf/iclr/DhingraZBNSC20} 
use BERT-like pretrained language model to encode the question and passages independently into dense representations, and use maximum inner product search (MIPS) algorithms~\cite{DBLP:conf/nips/Shrivastava014} to efficiently retrieve the most similar passage for each question. In this paper, we utilize Dense Passage Retriever (DPR)~\cite{DBLP:journals/corr/abs-2004-04906} as the base QA model.

\paragraph{Relation Bias of Existing QA Datasets. }\label{sec:rel}
We first explore how much relational knowledge between entities is required to answer the questions in the existing open-domain QA dataset. We conduct an empirical study to analyze the relation distribution in Natural Questions, one of the largest open-domain QA datasets, and how it influences QA model's performance.

For each question in Natural Question training set, we first select the entity 
that the ground-truth passage is associated with. We then combine the entity with the answer as an entity pair, 
and check whether we can find a relation triplet in WikiData describing the relation between these two entities.
Out of 58,880 training QA pairs, there are 23,499 pairs that could be aligned.
The aligned QA pairs cover 329 relations, which accounts for 16.4\% of the total 2,008 relations in WikiData. For most unaligned QA pairs, the answers are not entities and thus cannot be aligned to the graph. 

In addition to the low relation coverage issue in Natural Question, we also find that the relation distribution is imbalanced. As showed in Figure~\ref{fig:cdf}, 90\% of relations have frequency less than 41, and 30\% of relations appear only once. On the contrary, the most frequent relation ``P161 (cast member)'' appears 1,915 times out of 9,238 aligned QA pairs. A complete list of all these relations with aligned QA pairs is shown in Table~\ref{tab:freq1}-\ref{tab:freq2} in Appendix.

We then study whether the imbalanced relation distribution influences the performance of QA models trained on these datasets. We use a DPR model trained on training set of Natural Questions and then calculate the Exact Match accuracy in validation set
of each aligned QA pairs. 
We then analyze the correlation of the accuracy with the relation frequency in training set. As illustrated in Figure~\ref{fig:emp}, the validation set accuracy is overall proportional to the relation frequency in  training set. For those relations with frequency less than 5, the average accuracy is only 20.3\%, much lower than the average accuracy 42.7\% over all samples in validation set. 
This shows that the relation bias in existing QA datasets severely influences the generalization of QA models to questions with long-tail relations.



\section{Method}\label{sec:methodology}In this section, we will discuss \method framework in: 1) how to generate relational QA dataset for the pre-training purpose; and 2) how to construct a self-training task to empower QA model to capture relational facts.

\subsection{Construct QA Pre-Training Dataset}~\label{sec:qg}
To help QA model capture the knowledge from relation facts required to answer open-domain questions, we first focus on generating QA pre-training dataset, in which there exist relation connections between the source entity in questions to the target answer.
Specifically, each QA pair datapoint 
$d=\big\langle \langle s, r, t \rangle, q,p^{+} \big\rangle$ consists of three components: 1) relational triplet $\langle s, r, t \rangle$, in which $r$ denotes the relation between source entity $s$ and target entity $t$; 2) question $q$ in natural language asking which entity has relation $r$ to source entity s, with target entity $t$ as the correct answer;
3) positive context passage $p^{+} \in \mathbb{C}[s]$, a passage from source entity's Wiki-page that contains the target answer $t$. 


\paragraph{\kg. }  To generate QA pre-training dataset, leveraging the relation triplets in knowledge graph, e.g., WikiData, is a natural choice to define questions that require relation reasoning. 
We therefore construct \kg, in which each relation triplet $\langle s, r, t \rangle$ is linked to a set of description passages $\{desc.(s, t)\}$ in the Wiki-page of entity $s$. These descriptions would be later utilized to generate questions $q$ and positive context passages $p^{+}$.



To construct such a graph, we use the 2021 Jan. English dump of Wikidata and Wikipedia. For each Wikipedia hyperlink $\langle s, ?, t \rangle$ ($?$ denotes the relation is unlabeled), the passage containing anchored text to $t$ in the Wiki-page of $s$ naturally fits our requirement for $desc.(s,t)$. For each WikiData relation triplet $\langle s, r, t \rangle$, if the two entities are linked by a hyperlink in Wikipedia, we label the relation of the aligned hyperlink as $r$. For the other triplets $\langle s, r, t \rangle$ without alignment with hyperlinks, we extract all mentioning of target entity $t$ from the Wiki-page of $s$, and use the context passage as $desc.(s,t)$. The dataset statistics are shown in Table~\ref{tab:stat}. 

\begin{table}[t]
\small
\centering
\begin{tabular}{l|cc} \toprule
$\#$ of linked Entity & 5,640,366\\ 
$\#$ of relation labels & 2,008 \\ 
$\#$ of labelled triplet & 14,463,728  \\ 
$\#$ of unlabeled triplet (hyperlink) & 66,796,110 \\ 
$\#$ of grounded descriptions per triplet &  1.25\\
\bottomrule
\end{tabular}
\caption{Statistics of \kg.}
\label{tab:stat}
\end{table}

\begin{figure*}[t]
\begin{center}
\includegraphics[width=1.9\columnwidth]{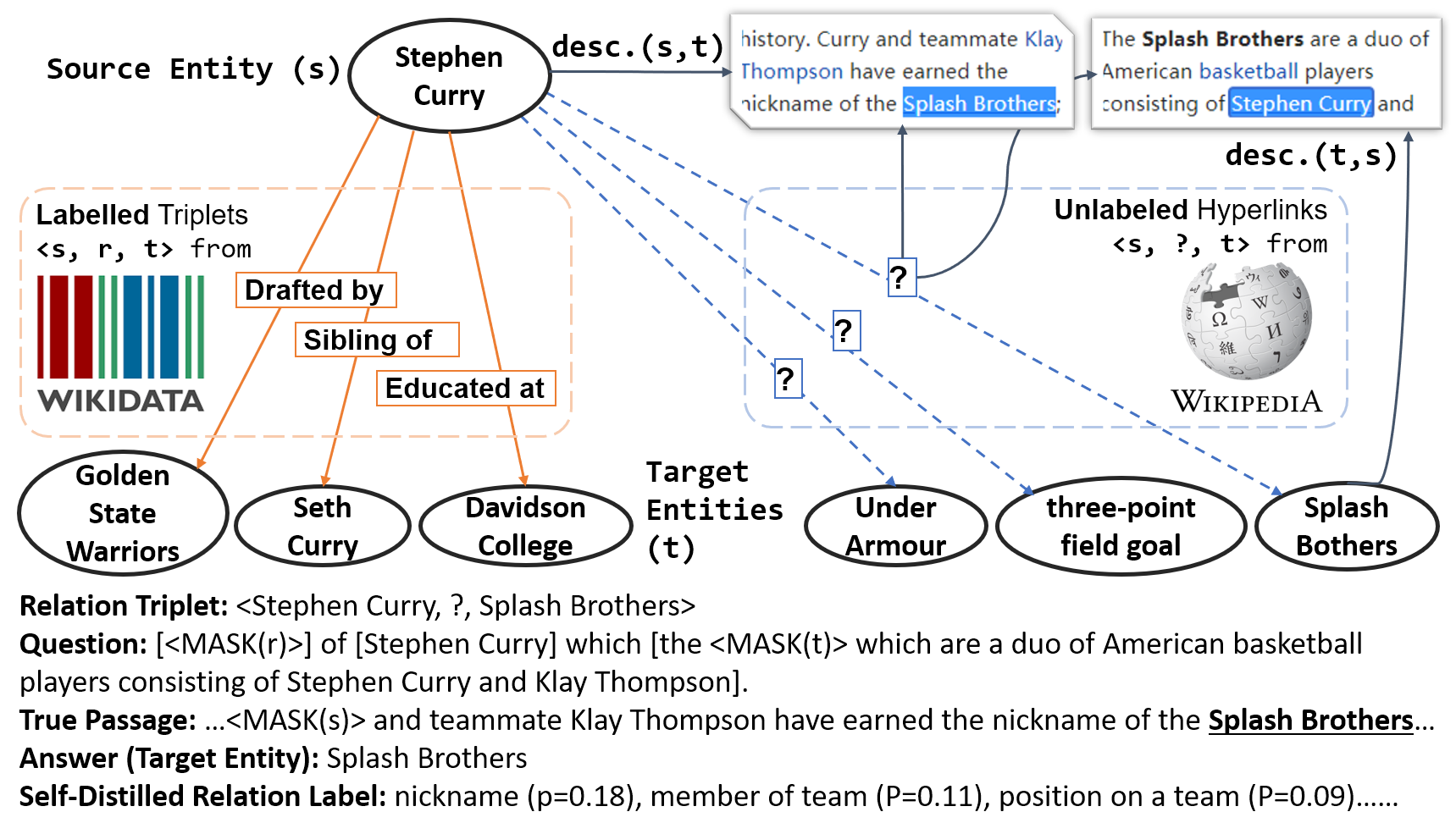}
\caption{Example of a generated relational QA pair from \kg.}\label{fig:example}
\end{center}
\end{figure*}

\paragraph{Relational QA Pair Generation}
In the following, we introduce the details to generate the relational QA pair from the constructed graph. 

Recent unsupervised QA studies~\cite{DBLP:conf/acl/LiWDWX20, DBLP:journals/corr/abs-2010-12623} revealed that if the question $q$ and context passage $p^{+}$ share a large lexical overlap, then the QA model could utilize low-level lexical patterns as shortcuts to find the answer. These shortcuts hinder the model from learning to comprehend the passages and answer the questions, hurting model's generalizability. To avoid this lexical overlap issue, we aim to generate questions from a passage that is different from the context passage $p^+$.

We first select all the entity pairs $\langle s, t \rangle$ that have mutual links in the \kg, with $desc.(s,t)$ and $desc (t,s)$ in part of Wikipage of $s$ and $t$ respectively
, describing the relationship between the two entities. Without loss of generality, we denote $s$ as source entity and $t$ as the target answer. The passage $desc.(s,t)$ containing target answer $t$ can be used as the positive passage $p^{+}$. 

Next, we generate a question that is lexically different from $p^{+}$ using the following template:
\begin{align*}
    q(s,r,t) = [\texttt{MASK}(r)]\ \text{of}\ [s] \text{ which } [desc.(t,s)]? 
\end{align*}
in which \texttt{MASK}$(r)$ is a relation mask token. As $desc.(t, s)$ contains source entity $s$, it provides information to describe the relationship between $s$ and $t$, based on which the QA model should learn to infer the latent relation $r$, and retrieve positive passage $p^{+}=desc.(s,t)$ and extract answer entity $t$. 
In addition, as $desc.(t, s)$ and $desc.(s, t)$ come from different Wiki-page, our question generation procedure can avoid the lexical overlap issue that often occur in prior Unsupervised QA methods.

\paragraph{Mask Target Answer.} As description $desc.(t, s)$ is from target answer $t$'s wiki-page, it often contains the name of entity $t$. We thus need to mask $t$ from the question. Otherwise, the pre-trained model can simply identify the answer to a question based on the local patterns.

As an example, in Figure~\ref{fig:example}, we show how to generate question for triplet $\langle \text{Stephen Curry}, ?, \text{Splash Brothers} \rangle$. We firstly retrieve two descriptive passages $desc.(s,t)$ and $desc.(t,s)$ in two entities' wiki pages. Using the template, we generate the question along with the ground-truth passage. We then mask out the target entity in question and source entity in true passage (will discuss later in retrieval pre-training) to avoid shortcut. A list of generated relational QA pairs are shown in Table~\ref{tab:example} in Appendix.


\subsection{Relation-Guided QA Pre-Training}
With the generated relational QA dataset, we introduce how to pre-train both retriever and reader components in the QA model.

\subsubsection{Relation Prediction Pre-Training}
Our generated QA dataset contains the relation label $r$ between the source entity $s$ and the answer target $t$. Therefore, we design a self-training task to guide the model to predict the latent relation in question, which can benefit both retriever and reader. Specifically, we adopt a linear projection layer $\text{L}_{R}(\cdot)$ over the $\texttt{BERT}_{\texttt{[CLS]}}$ token embedding to predict the relation over the WikiData relation set. The pre-training loss of relation prediction is:
\begin{align}
    \mathcal{L}_{\text{rel}} = \frac{1}{B}\sum_{q} -\log P(r \mid q ; \theta), \nonumber
\end{align}

\paragraph{Self-Distillation for Unlabelled Relation}
The hyperlinks in wikipedia also provide valuable implicit information about the relations between entities. 
To leverage them, we use the trained relation predictor at each epoch with fixed parameter $\hat{\theta}$ as teacher model to assign soft label and then progressively train the relation predictor as student model based on the assigned labels in the next epoch. This approach is referred to as self-distillation in the literature~\cite{DBLP:conf/cvpr/XieLHL20, DBLP:conf/nips/ChenKSNH20}. 
We minimize this self-distillation loss as:
\begin{align}
    \mathcal{L}_{\text{distill}} \!=\! \frac{1}{B}\sum_{q} \sum_{\hat{r}}  -\log P(\hat{r} \!\mid\! q ; \theta) \cdot \text{sg}\big(P(\hat{r} \!\mid\! q ; \hat{\theta})\big), \nonumber
\end{align}
where sg$(\cdot)$ denotes the operation of stop gradient, which avoids back propagation to the teacher network with fixed parameter $\hat{\theta}$. $\hat{r}$ is enumerating all the relation labels.

As the relation predictor at early stages cannot give a reasonable prediction, we put a dynamic weight schedule to 
$\mathcal{L}_{\text{distill}}$ by a time-dependent weighting term $1-e^{-\text{epoch}}$, which ramps up from zero to one. Combing the weighted self-distillation loss $\mathcal{L}_{\text{distill}}$ with the supervised relation loss $\mathcal{L}_{\text{rel}}$, 
we get the final relation loss $\hat{\mathcal{L}}_{\text{rel}}$ to train the model capturing all relational facts covered in the \kg. 

\subsubsection{Dense Retrieval Pre-Training}

The goal of dense retrieval pre-training is to get a question encoder $\text{Enc}_Q$ and a passage encoder $\text{Enc}_P$ to map questions and all passages in the Wiki Corpus $\mathbb{C}$ into an embedding space, such that each question $q$ is close to its ground-truth positive context passage $p^{+}$ in the embedding space. 
The objective is as follows: 
\begin{align} \label{eq:retr_prob}
P_{\text{retr}} (p^{+} \mid q,\mathbb{C}) =  \frac{\exp\big(\text{sim}(q, p^{+})\big)}{\sum_{p \in \mathbb{C}} \exp\big(\text{sim}(q, p)\big)},
\end{align} 
where $\text{sim}(q, p)$ is the cosine similarity between the normalized embeddings of question and passage. 



\paragraph{Two-Level Negative Passage Sampling.} \label{inbatch}
As we cannot enumerate all other passages in the denominator of Eq(\ref{eq:retr_prob}), we need to sample a set of negative passages for contrastive learning.
Previous studies~\cite{DBLP:journals/corr/abs-2004-04906} have revealed that it is essential that the sampled negative passages should be hard enough to train the retriever.
As the question and passage embeddings are encoded independently, DPR can efficiently calculate the similarity of each question to all passages in the batch via dot product. Based on this property, as long as the passages within a batch are similar to each other, they serve the hard cases of negative passages to others. We thus propose a two-level negative passage sampling strategy to construct hard cases for training the retriever in the following.

We first sample at the level of entity. Given a set of randomly sampled $b$ entities, we adopt random walk from these seed entities over the \kg to get $B$ entities. As the connected entities have a relationship, their true passages are also semantically similar, and thus serve as good negative samples. We then conduct sampling at the level of passage. For each source entity $s_i$ with positive passage $p^{+}_i \in \mathbb{C}[s_i]$, we randomly pick $K$ other passages from the same Wiki-page to form a negative passage set $\big\{p^{-}_{i,j} \in \mathbb{C}[s_i], \text{s.t.}\ p^{-}_{i,j} \neq p^{+}_i\big\}_{j=1}^{K}$. These negative passages are similar to $p^{+}_i$, as they all describe the same entity $s_i$. 


After we collect both the positive and $K$ negative passages for all the entities, we use the passage encoder $\text{Enc}_P$ to get a passage embedding matrix $\mathbf{P}$ with dimension $\big((1+K)\cdot B \times d\big)$. We also use question encoder $\text{Enc}_Q$ to get question embedding matrix $\mathbf{Q}$ with dimension $\big(B \times d\big)$. We then get a similarity matrix $\mathbf{S} = \mathbf{Q} \mathbf{P}^{T}$ with dimension $\big(B \times (1+K)\cdot B\big)$, in which the diagonal entry corresponds to the similarity between question and its positive passage. 
We thus calculate the retrieval loss with in-batch negative samples via:
\begin{align}
    \mathcal{L}_{\text{retr}} = \frac{1}{B} \Big( \sum_{i\in[1,B]}\big(-\log \text{softmax}(\mathbf{S})\big)_{[i,i]} \Big).
\end{align}

\paragraph{Masking Source Entity.} As the true passage $p^{+}_i=desc.(s,t)$ might contain the name of source entity $s$. We mask out all the tokens of $s$ from the extracted passages, so that the model is required to understand the passages for correct retrieval instead of exploiting a shortcut. 


\subsubsection{Reading Comprehension Pre-Training}

The goal of reading comprehension pre-training is to get a neural reader that re-ranks the top-$k$ retrieved passages and extracts an answer span from each passage as the answer. The probability of a passage contains the target answer $t$, and each token in the selected passage being the starting/ending positions of an $t$ are defined as:
\begin{align}
    &P_{\text{rank}}(t \in p) = \frac{\exp\big(\text{L}_{\text{rank}}\big(\texttt{BERT}_{\texttt{CLS}}(q, p)\big)\big)}{\sum_{\hat{p}} \exp\big(\text{L}_{\text{rank}}\big(\texttt{BERT}_{\texttt{CLS}}(q, \hat{p})\big)\big) }, \nonumber \\
    &P_{\text{start}}(i \mid p, q) = \frac{\exp\big(  \text{L}_{\text{start}}  \big(\texttt{BERT}_{\texttt{[i]}}(q, p) \big) \big)} {\sum_{j} \exp\big(  \text{L}_{\text{start}}  \big(\texttt{BERT}_{\texttt{[j]}}(q, p) \big) \big)}, \nonumber \\
    &P_{\text{end}}(i \mid p, q) = \frac{\exp\big(  \text{L}_{\text{end}}  \big(\texttt{BERT}_{\texttt{[i]}}(q, p) \big) \big)} {\sum_{j} \exp\big(  \text{L}_{\text{end}}  \big(\texttt{BERT}_{\texttt{[j]}}(q, p) \big) \big)}. \nonumber
\end{align}
where L$_{*}$ are linear project layers with different parameters. Note that the re-ranking module adopts cross-attention over questions and passages rather than the dot product of two independently encoded embedding used in retriever. 
For each QA pair $d=\big\langle \langle s, r, t \rangle, q,p^{+} \big\rangle$, 
we select $m$ other passages in wiki-page of entity $s$ as negative passages, and maximize $P_{\text{rank}}(t \in p^{+})$. Then, we calculate $P_{\text{start}}(i \mid p^{+}, q)$ and $P_{\text{end}}(i \mid p^{+}, q)$ and maximize the probability for the ground-truth span of target answer $t$. Combing the passage re-ranking and span extraction objectives, we get reading-comprehension loss $\mathcal{L}_{\text{read}}$.

\section{Experiments}\label{sec:experiment}In this section, we evaluate \method\ on three open-domain QA datasets: Natural Questions (NQ), Trivia QA and Web Questions (WQ). 

\subsection{Experiment Settings}




We follow the pre-processing procedure described in DPR~\cite{DBLP:journals/corr/abs-2004-04906} for a fair comparison. We use the English Wikipedia from Dec. 20, 2018 and split each article into passages of 100 disjoint words as the corpus. For each question in all the three datasets, we use a passage from the processed Wikipedia which contains the answer as positive passages. We evaluate the QA system by Exact Match (EM) Accuracy on the correct answer.

Our \method could be integrated with any open-domain QA system. In this paper, we incorporate it with the recently developed QA system, Dense Passage Retriever (DPR)~\cite{DBLP:journals/corr/abs-2004-04906} to evaluate our pre-training framework. The DPR model uses the RoBERTa-base (d=768, l=12) model as the base encoder.
We first pre-train the retriever and reader in DPR using \method. For retriever, we use the negative passage sampling strategy (c.f. Sec.~\ref{inbatch}), with initial entity size set to be 12, batch size of 128 and the hard negative passage number of 2. For reader, we randomly sample 64 source entities per batch to calculate the loss. For each entity, we sample 2 hard negative passages for re-ranking. We pre-train both the retriever and reader for 20 epochs using AdamW optimizer and a learning rate warm-up followed by linear decay. Pre-training is run on 8 Tesla V100 GPUs for two days.
After the pre-training, we fine-tune the retriever and reader on each QA dataset following the same procedure and hyper-parameters described in DPR~\cite{DBLP:journals/corr/abs-2004-04906}. 

\begin{table*}[ht]
\centering
\small
\begin{tabular}{ll|c|ccc} \toprule
& \multirow{2}{*}{\textbf{QA System Name}}    &  \textbf{Pre-Training} & \textbf{NQ} & \textbf{Trivia QA} & \textbf{WQ}\\ 
& & \textbf{Task for QA}& (58.9k/3.6k) & (60.4k/11.3k) & (2.5k/2k)\\ \midrule
\multirow{5}{*}{\rotatebox[origin=c]{90}{{Supervised}}} & BM25+BERT~\cite{DBLP:conf/acl/LeeCT19}   & -           & 26.5  &   47.1   &   17.7  \\
~& HardEM~\cite{DBLP:conf/emnlp/MinCHZ19}  & -           & 28.1  &   50.9   &     -   \\
~& GraphRetriever~\cite{DBLP:journals/corr/abs-1911-03868}    & -     & 34.5  &   56.0   &   36.4  \\ 
~& PathRetriever~\cite{DBLP:conf/iclr/AsaiHHSX20}    & -      & 32.6  &     -    &     -   \\
~& DPR~\cite{DBLP:journals/corr/abs-2004-04906}  & - & 41.5  &   56.8   &   34.6  \\ \midrule
\multirow{8}{*}{\rotatebox[origin=c]{90}{{Pre-Trained for QA}}} &T5 (large) ~\cite{DBLP:journals/jmlr/RaffelSRLNMZLL20} & T5 (Multitask)  & 29.8 & - & 32.2 \\
~& ORQA~\cite{DBLP:conf/acl/LeeCT19}       & ICT         & 33.3  &   45.0   &   36.4  \\ 
~& REALM$_{\text{Wiki}}$~\cite{DBLP:journals/corr/abs-2002-08909} & REALM & 39.2  &     -    &   40.2  \\ 
~& REALM$_{\text{News}}$~\cite{DBLP:journals/corr/abs-2002-08909} & REALM & 40.4  &     -    &   40.7  \\ 
~& DPR (KnowBERT~\cite{DBLP:conf/emnlp/PetersNLSJSS19})  & Entity Linking  & 39.1  &   56.4   &   34.8  \\ 
~& DPR (KEPLER~\cite{DBLP:journals/corr/abs-1911-06136})  & TransE  & 40.9  &   57.1   &   35.2  \\
~& DPR (Unsup.QA~\cite{DBLP:conf/acl/LewisDR19}) & Cloze Translation & 41.9 & 57.3 & 36.5 \\   \cmidrule{2-6}
~ & Ours, DPR (\method)   &  \method               & \textbf{43.7}  &   \textbf{59.2}   &   \textbf{40.9}  \\ \bottomrule
\end{tabular}
\caption{\textbf{End-to-end QA} Exact Match Accuracy (\%) on test sets of three Open-Domain QA datasets, with the number of train/test examples shown in paretheses below. All the results except the last four rows are copied from the original papers. “–” denotes no results are available. Models in the first block are initialized by BERT/RoBERTa and then directly fine-tuned on the supervised QA datasets. While models in the second block are initialized by RoBERTa and then tuned on some QA pre-training tasks first, and then fine-tuned on the supervised QA datasets.}
\label{tab:e2e}
\end{table*}

\paragraph{QA Pre-Training Baselines.}
We compare \method with three recently proposed pre-training methods for open-domain QA.

\textbf{T5}~\cite{DBLP:journals/jmlr/RaffelSRLNMZLL20} adopts multiple generative tasks to pre-train a generative model. The fine-tuned QA models directly generate answers 
without needing an additional retrieval step.

\textbf{ORQA}~\cite{DBLP:conf/acl/LeeCT19} adopts a Inverse Cloze Task (ICT) to pre-train retriever, which forces each sentence's embedding close to context sentences.

\textbf{REALM}~\cite{DBLP:journals/corr/abs-2002-08909} incorporates a retriever as a module into language model and trains  the whole model over masked entity spans.

We directly report the results listed in their papers as they follow the same experiment settings. 

We also add two knowledge-guided language models as baselines. Though not targeted at QA problem, these two methods are both designed to capture structured knowledge.

\textbf{KnowBERT}~\cite{DBLP:conf/emnlp/PetersNLSJSS19} adds entity embedding to each entity mention in text, and adopts the entity linking objective to pre-train the model.

\textbf{KEPLER}~\cite{DBLP:journals/corr/abs-1911-06136} uses Knowledge Embedding objective, i.e., TransE, to guide embedding encoded over entity description.

We initialize DPR base encoders by the released pre-trained models of these two work, and then fine-tune on each QA dataset with the same procedure.

We also add a Unsupervised Question Answering (\textbf{Unsup.QA})~\cite{DBLP:conf/acl/LewisDR19} as a baseline. 
For each entity as the answer, Unsup.QA selects a passage containing the entity as context passage and a cloze question.
The cloze question is later re-written by a machine translator to natural language. We use the generated QA dataset to pre-train both the retriever and reader of the DPR framework.

\subsection{Experimental Results} 



\begin{table}[ht]
\centering
\small
\begin{adjustwidth}{-2mm}{}
\begin{tabular}{l|ccc} \toprule
Pre-Train Model     & NQ & Trivia QA & WQ\\  \midrule
RoBERTa             & 78.4 / 63.3  &   79.4 / 72.6   &   73.2 / 58.1  \\
KnowBERT  & 76.7 / 62.6  &   78.9 / 72.2   &   73.4 / 58.3  \\
KEPLER  & 77.9 / 62.8  &   79.7 / 72.9   &  74.5 / 58.6   \\
Unsup.QA &  78.6 / 63.7  &   79.9 / 73.0   &  74.5 / 59.1  \\ \midrule
\method                      & \textbf{80.1} / \textbf{64.8}  &  \textbf{81.2} / \textbf{73.7}   &   \textbf{76.7} / \textbf{61.0}  \\ 
\bottomrule
\end{tabular}
\end{adjustwidth}
\caption{\textbf{Retrieval (left)} accuracy over Top-20 results and \textbf{Reader (right)} Exact Match over Golden-Passages on validation sets of three Open-Domain QA datasets. }
\label{tab:read}
\end{table}


\begin{table}[ht]
\centering
\small
\begin{adjustwidth}{-2mm}{}
\begin{tabular}{cccc|ccc} \toprule
Mask & NPS  &  $\mathcal{L}_{\text{distill}}$    & $\mathcal{L}_{\text{rel}}$    & NQ &  Trivia QA & WQ \\ \midrule
\cmark &\cmark & \cmark & \cmark &  44.3 & 59.8 & 41.4  \\
\xmark & \cmark &\cmark & \cmark &  39.7 & 56.3 & 34.2  \\
\cmark & \xmark &\cmark & \cmark &  43.5 & 58.1 & 39.8  \\
\cmark & \cmark &\xmark & \cmark &  43.8 & 59.3 & 40.8  \\
\cmark & \cmark &\xmark & \xmark &  43.1 & 58.5 & 40.0  \\ \bottomrule
\end{tabular}
\caption{\textbf{Ablation} of \method components on validation sets of three Open-Domain QA datasets. Mask: Mask target entity from question and source entity from passage; NPS: Two-level Negative Passage Sampling. }
\end{adjustwidth}
\label{tab:ablation}
\end{table}

\begin{table}[ht]
\centering
\small
\begin{tabular}{cc|ccc} \toprule
B & K   & NQ &  Trivia QA & WQ \\ \midrule
128  &   2    &  80.1     & 81.2    &  76.6 \\
128  & 1   &79.7     & 80.8   & 76.1 \\
64   &        2         &79.6     &  80.6      & 75.8 \\
64   &  1    &79.2    &  80.1       &75.3 \\ \bottomrule
\end{tabular}
\caption{\textbf{Ablation} of batch size and negative sampling for retrieval pre-training. B: Batch Size; K: Number of other passages as negative sample.}
\label{tab:neg}
\end{table}

Table~\ref{tab:e2e} summarizes the overall EM accuracy of the QA systems on the three datasets. The DPR framework pre-trained by \method outperforms all other open-domain QA systems. Comparing with DPR without pre-training, \method achieves 2.2\%, 2.4\% and 6.3\% enhancement in EM accuracy on the three datasets. 


Comparing with other pre-training tasks for QA, \method outperforms ORQA by 10.4\%, 14.2\% and 4.5\% on the three datasets, and outperforms REALM$_\text{News}$ by 3.3\% and 0.2\% on NQ and WQ. This demonstrates that the model performance can be enhanced by leveraging relational QA dataset guided by \kg. We provide a detailed analysis in Sec.~\ref{sec:generalize}.



KnowBERT and KEPER encode structural knowledge into pre-trained language models. Both models focus on generating meaningful entity embedding, and are not designed to infer relations between entities for question answering. From the table, KEPLER trained via TransE performs slightly better than KnowBERT trained via entity linking, and \method outperforms KEPLER by 2.8\%, 2.1\%, 5.7\% on the three datasets.

Similar to \method, Unsup.QA~\cite{DBLP:conf/acl/LewisDR19} also generates QA data from Wikipedia. This baseline slightly improves DPR by 0.4\%, 0.5\%, 1.9\% on the three datasets, while our \method outperforms it by 1.8\%, 1.9\%, 4.4\%. As discussed in Sec~\ref{sec:qg}, one of the main reasons that our graph-based QA generation strategy performs better is that we adopt grounded description passages $desc.(t,s)$ and $desc.(s,t)$ from different documents as questions and contexts. This avoids the lexical overlap problem in Unsup.QA and help model to capture relational facts.


We also show the retrieval and reader performance separately on validation sets in Table \ref{tab:read}. Compared with DPR without pre-training, \method improves top-20 accuracy of Retriever by 1.7\%, 1.8\%, and 3.5\%, and improves EM accuracy of Reader by 1.5\%, 1.1\%, and 2.9\%. Also, \method outperforms all the other pre-training baselines. This shows that \method improves both the retrieval and reader steps of open-domain QA.

\paragraph{Ablation Studies. } We then analyze the importance of each model component in \method. 
One key strategy is to mask out the target answer from questions and mask out source entities from passages during retrieval training. This can avoid the model using the entity surface to find the correct passage and answer. Without using masking strategy, the average EM performance drops 5.1\%. This shows that
it is essential to apply the mask strategy to avoid shortcut in QA pre-training. Next, we replace the hard negative passage sampling during retrieval pre-training with random batch sampling. The average EM performance drops 1.4\%, showing the importance of hard negative samples. Finally, we study the unsupervised relation loss $\mathcal{L}_{\text{distill}}$  and the supervised $\mathcal{L}_{\text{rel}}$. Removing them leads to 0.5\% and 1.3\% performance drop, which shows the benefit of training the model to explicitly infer the relation from questions.

Another key component is the negative passage sampling for dense retrieval pre-training. We study how the batch size and number of negative sample influence the performance of trained retrieval. As is shown in Table~\ref{tab:neg}, increasing batch size and negative sample size can improve the performance of retriever. Even with a small batch size and negative sample, our pre-training framework could still achieves better performance against non-pretrain baseline, showing that our approach is not sensitive to these two hyperparameters.

\begin{figure}[t]
\begin{center}
\includegraphics[width=1.0\columnwidth]{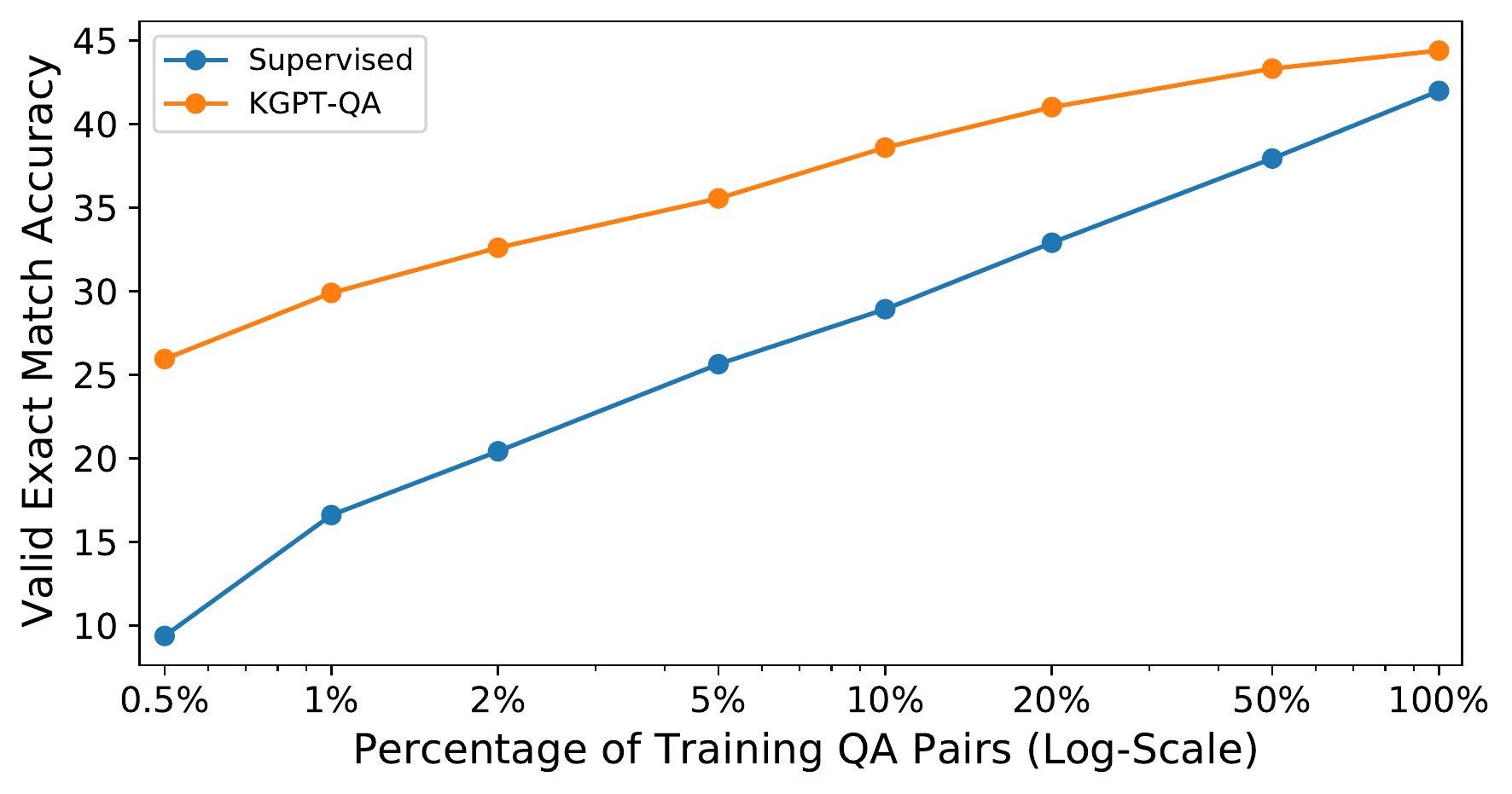}
\caption{Few-shot QA experiment. Figure shows EM accuracy in validation set of DPR model with and without \method pre-training, fine-tuned with different percentage of data on Natural Questions.}\label{fig:few}
\end{center}
\end{figure}

\begin{figure}[ht]
\begin{center}
\includegraphics[width=1.0\columnwidth]{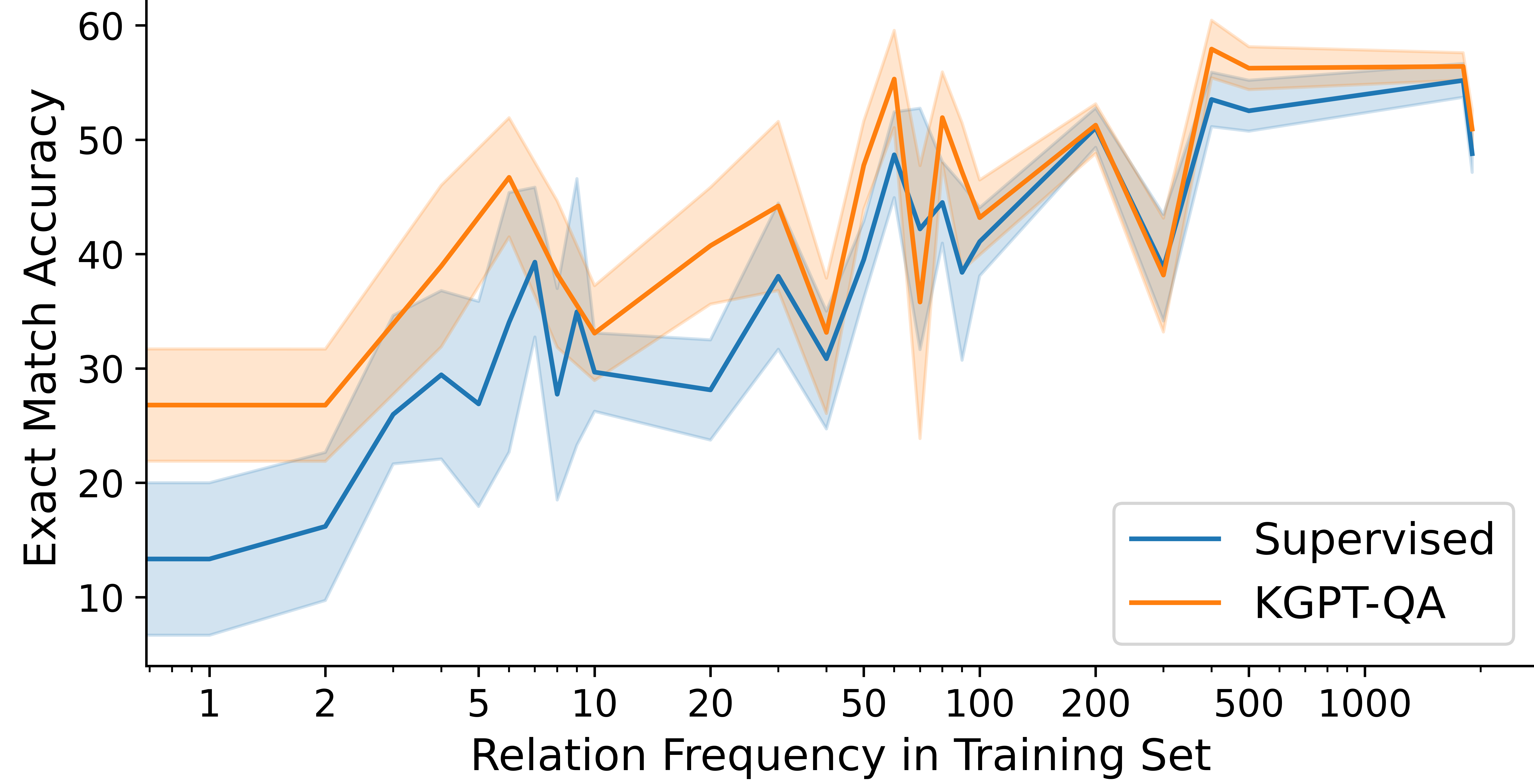}
\caption{Long-tail relation experiment. EM accuracy of questions in validation set with different relation frequency in training set.}\label{fig:relation}
\end{center}
\end{figure}

\paragraph{Few-Shot QA Performance. }We analyze the improvement of \method when only a few labelled training samples are available. We fine-tune DPR initialized by \method on subset of Natural Questions with different percentages. As is shown in Figure~\ref{fig:few}, \method consistently outperforms DPR without pre-training, and the improvement is more significant with small data. Specifically, when only 0.5\% (594) labelled QA pairs are provided, the DPR pre-trained by \method can still achieve 26.0\% Val EM accuracy, significantly higher than 9.4\% achieved by the DPR without pre-training. The results show that \method provides a good initialization for QA systems and reduce the requirement of large human-annotated QA dataset.

\subsection{Generalization for long-tail relations. }\label{sec:generalize}
As pointed out in Section~\ref{sec:rel}, existing QA datasets suffer high relation bias, and thus a QA model trained on these datasets cannot generalize well to questions with long-tail relations. We thus analyze whether our \method can remedy this issue. As is shown in Figure~\ref{fig:relation}, the performance improvement of \method against the supervised baseline is much more significant for the questions with infrequent relations. Specifically, for all relations appear less than 5 times in training set, the average EM accuracy of \method is 33.3\%, significantly higher than 22.4\% achieved by DPR without pre-training. This indicates that our relation QA generation method could indeed improve the performance on QA pairs with long-tail relations. Detailed prediction results are shown in Table~\ref{tab:compare} in Appendix.

\section{Related Works}\label{sec:related}\paragraph{Unsupervised QA via Question Generation} To train a QA system without human annotation of QA pairs, Unsupervised QA has been proposed by \citet{DBLP:conf/acl/LewisDR19} to generate synthetic $\langle context, question, answer \rangle$ data for training QA models. 
\citet{DBLP:conf/acl/LewisDR19} synthesize the QA data by: 1) run NER or noun chunkers over randomly sampled English Wikipedia paragraphs to extract $answers$; 2) Treat the paragraphs surrounding the answer as $context$; 3) Treat the context as cloze\-style question and feed into a unsupervised machine translator to generate $natural\  questions$. Some follow-up works also utilize template~\cite{DBLP:conf/acl/FabbriNWNX20} and pre-trained language model~\cite{DBLP:conf/emnlp/PuriSSPC20} over masked cloze-style questions for more human-readable questions.
These cloze-style unsupervised QA methods achieve promising performance than previous heuristic QA baselines but underperform supervised ones. The main limitation is that the question is generated with the masked context as input, resulting in severe overlap of lexicon and word surface with the context. Consequently, the QA model might utilize the lexical pattern as a shortcut to find the answer. 
To address the problem of context-question lexical overlap, \citet{DBLP:conf/naacl/DhingraPR18} assume each article has an introductory paragraph, and use this paragraph to generate answer. \citet{DBLP:conf/acl/LiWDWX20} retrieve the Wikipedia cited document as context, \citet{DBLP:journals/corr/abs-2010-12623} leverage structured tables to extract key information from context, with which to synthesize questions. 

To tackle the challenges in previous studies, our framework propose to leverage the Wikipedia hyperlinks and Wikidata relations as the bridge to connect two entities with linked descriptions. With one description as question and the other as context, the question and context are semantically relevant and lexical different, which naturally solve the problem without involving any additional module. 

\paragraph{Knowledge-Guided Pre-Training}
Recently, researchers investigated to inject structured knowledge into pre-trained language models. ~\citet{DBLP:conf/acl/ZhangHLJSL19} and ~\citet{DBLP:conf/emnlp/PetersNLSJSS19} propose to add entity embedding to each entity mentions in text, and add entity linking objective to guide model capture structured knowledge. \citet{DBLP:journals/corr/abs-1911-06136} encode entity text description as entity embeddings and train them via TransE objective. Though these work show improvements over several natural language understanding tasks, they are not dedicated to open-domain question answering tasks.

There are also several pre-training studies for QA. For retrieval, \citet{DBLP:conf/acl/LeeCT19} propose an inverse cloze task, which treats a random sentence as query and the surrounding contexts as ground-truth evidence to train a QA retrieval model. \citet{DBLP:conf/icml/GuuLTPC20} propose to explicitly add a retriever module in the language model to train the retriever via language modelling pre-training. For reader, \citet{DBLP:conf/iclr/XiongDWS20} propose to a weakly supervised pre-training objective. They construct some fake sentences by replacing the entities in a sentence with the other entities of the same type, and train the model to discriminate original sentence from the fake ones. \citet{DBLP:journals/corr/abs-2007-00849} incorporate the knowledge graph triplets into language model, so the model could utilize the triplets to predict correct entity. \citet{DBLP:journals/corr/abs-2102-07043} extend this work by learning a virtual knowledge base by inferring the relation between two co-occurring entity pairs. 

Compared with these works, our \method mainly differs in: 1) We do not change the base QA model, so the pre-training framework could be applied to any QA systems. 2) We explicitly model the relations between entities, which proves to benefit QA pairs with less frequent relation patterns.
\section{Conclusion}\label{sec:conclusion}In this paper, we propose a simple yet effective pre-training framework \method. We leverage both the Wikipedia hyperlinks and Wikidata relation triplets to construct \kg, based on which we generate relational QA dataset. 
We then pre-train a QA model to infer the latent relation from the question, and then conduct extractive QA to get the target answer entity. \method improves the performance of the state-of-the-art QA frameworks, especially for questions with long-tail relations.
\section*{Acknowledgement}
This work was partially supported by NSF III-1705169, NSF 1937599, DARPA HR00112090027, Okawa Foundation Grant, and Amazon Research Awards.

\bibliography{emnlp}
\bibliographystyle{acl_natbib}

\newpage
\appendix

\clearpage

\begin{table*}[ht]
\scriptsize
\centering
\begin{adjustwidth}{-5mm}{}
\begin{tabular}{l|c|l|l} \toprule
 \textbf{Relation}    & \textbf{Frequency} & \textbf{Question} & \textbf{True Answer}\\ \midrule
P161 (cast member)&1915&what was the geeks name in 16 candles&anthony michael hall\\ 
P175 (performer)&1844&who sang the original blinded by the light&bruce springsteen\\ 
P676 (lyrics by)&519&who sings the song i can see clearly now the rain is gone&johnny nash\\ 
P86 (composer)&442&who made the beavis and butthead theme song&mike judge\\ 
P725 (voice actor)&334&who plays the voice of tiana in princess and the frog&anika noni rose\\ 
P1346 (winner)&283&who has won the 2017 womens singles wimbledon tennis tournament&garbiñe muguruza\\ 
P50 (author)&263&where does the saying standing on the shoulders of giants come from&bernard of chartres\\ 
P17 (country)&257&where did the black panther party take place&united states\\ 
P527 (has part)&198&the unit of area in mks system is&metre\\ 
P162 (producer)&134&who is in the video do n 't worry be happy&bobby mcferrin\\ 
P276 (location)&117&where will the summer olympics be held in 2020&tokyo\\ 
P840 (narrative location)&103&what state is a christmas story based in&indiana\\ 
P915 (filming location)&98&where was the movie the english patient filmed&tunisia\\ 
P710 (participant)&88&who died at the gunfight at okay corral&billy clanton\\ 
P170 (creator)&87&who came up with britain 's got talent&simon cowell\\ 
P1308 (officeholder)&87&who is the first lady of the usa&melania trump\\ 
P361 (part of)&74&who sings if you want to destroy my sweater&weezer\\ 
P39$_r$ (R: position held)&64&who is the attorney general for new jersey&gurbir grewal\\ 
P138 (named after)&64&who proved that mar 's orbit is elliptical not circular&nicolaus copernicus\\ 
P112 (founded by)&61&who created a settlement house with the help of other social reformers&ellen gates starr\\ 
P161$_r$ (R: cast member)&60&who is miss sue in the blind side&kathy bates\\ 
P31$_r$ (R: instance of)&57&the world 's oldest epic tale told in poetry is called the epic of&epic of gilgamesh\\ 
P58 (screenwriter)&57&who wrote the story for the shape of water&vanessa taylor\\ 
P61 (discoverer or inventor)&55&who developed the analytical engine which had features of present day computers&charles babbage\\ 
P26 (spouse)&53&who does young catherine marry in wuthering heights&hareton earnshaw\\ 
P1923 (participating team)&52&who did the bengals play in the super bowl&san francisco 49ers\\ 
P166$_r$ (R: award received)&51&which indian actor has won the most national awards&amitabh bachchan\\ 
P674 (characters)&50&who said better to reign in hell than serve in heaven&satan\\ 
P279 (subclass of)&49&when does dna replication occur during the eukaryotic cell cycle&mitosis\\ 
P361$_r$ (R: part of)&49&where does the transmission of electrical impulses in the heart begin&sinoatrial node\\ 
P131 (is located in)&46&where is saba university school of medicine located&saba\\ 
P279$_r$ (R: subclass of)&45&what are the names of the three pedals on a piano&soft pedal\\ 
P54 (member of sports team)&41&what team does steph curry brother play for&dallas mavericks\\ 
P1344$_r$ (R: participant in)&38&who won rupauls drag race all stars three&trixie mattel\\ 
P495 (country of origin)&37&where was the movie snow white and the huntsman filmed&united kingdom\\ 
P39 (position held)&34&who is the present speaker of lok sabha 2018&sumitra mahajan\\ 
P127 (owned by)&33&who owns the independent newspaper in the uk&alexander lebedev\\ 
P607$_r$ (R: conflict)&32&in the civil war who had more soldiers&union army\\ 
P31 (instance of)&32&what kind of bridge is the mackinac bridge&suspension bridge\\ 
P1441$_r$ (R: present in work)&29&what is the dads name in the adams family&gomez addams\\ 
P175$_r$ (R: performer)&28&who does sean astin play in lord of the rings&samwise gamgee\\ 
P36 (capital)&28&what is the capital of dadra and nagar haveli&silvassa\\ 
P921 (main subject)&24&what disease did susannah have in brain on fire&anti-nmda receptor encephalitis\\ 
P186 (material used)&22&what is the liquid in a magic 8 ball&alcohol\\ 
P179$_r$ (R: part of the series)&22&what is the second book in the mortal instruments series&city of ashes\\ 
P793$_r$ (R: significant event)&21&which territories did the us gain in the spanish-american war&puerto rico\\ 
P115 (home venue)&21&where does portland 's nba basketball team the portland trailblazers play&moda center\\ 
P371 (presenter)&21&who won beat bobby flay shrimp and grits&bobby flay\\ 
P180 (depicts)&19&who r the 4 presidents on mt . rushmore&abraham lincoln\\ 
P800$_r$ (R: notable work)&19&the explorer accurately mapped the coasts of europe and north africa&piri reis\\ 
P136 (genre)&17&scott joplin is best known as a composer of what kind of music&ragtime\\ 
P1431 (executive producer)&16&who hosted the daily show before trevor noah&jon stewart\\ 
P47 (shares border with)&16&which indian states share a border with delhi&uttar pradesh\\ 
P54$_r$ (R: member of sports team)&16&who scored the first goal in dallas stars history&neal broten\\ 
P144 (based on)&16&the tribute money depicts a scene from the&gospel of matthew\\ 
P57 (director)&15&who is the director of welcome to new york&chakri toleti\\ 
P488 (chairperson)&15&who is the leader of the democratic party now&tom perez\\ 
P403 (watercourse outflow)&15&what sea does the nile river flow into&mediterranean sea\\ 
P1889 (different from)&14&how to do alt codes on a mac&option key\\ 
P1441 (present in work)&14&when does luke skywalker find out leia is his sister&return of the jedi\\ 
P734 (family name)&14&who threw the first brick in the stonewall riots&johnson\\ 
P1269 (facet of)&14&which supreme court case established the separate but equal doctrine&plessy v. ferguson\\ 
P706 (takes place in)&13&what region of the world is greece in&southern europe\\ 
P176 (manufacturer)&13&who built the gerald r ford aircraft carrier&newport news shipbuilding\\ 
P84 (architect)&12&scottish architect who developed st martins in the field&james gibbs\\ 
P150 (contains)&12&what is the name of capital of argentina&buenos aires\\ 
P1532 (country for sport)&12&cristiano ronaldo what country does he play for&portugal\\ 
P800 (notable work)&12&what was the first book that charles dickens published&the pickwick papers\\ 
P641 (sport)&11&what is the number 1 sport in the usa&american football\\ 
P1001 (applies to jurisdiction)&11&who won the schenck v. united states case&united states\\ 
P206 (on lake)&11&where is ellis island located in new york&upper new york bay\\ 
P178 (developer)&11&ms office 2000 was developed by which company&microsoft\\ 
P166 (award received)&11&who won best actor in the academy awards this year&gary oldman\\ 
P102$_r$ (R: party)&11&who was known as the father of indian national congress&mahatma gandhi\\ 
P449 (original broadcaster)&10&what cbs channel is the late late show on&cbs\\ 
P2438 (narrator)&10&whos the main character in the great gatsby&nick carraway\\ 
P264 (record label)&10&who did the soundtrack for beverly hills cop&mca records\\ 
P674$_r$ (R: characters)&10&where is the story of joseph in the bible found&book of genesis\\ 
P1891 (signatory)&10&who has started reducing emissions from deforestation and forest degradation&brazil\\ 
P138$_r$ (R: named after)&10&roman god of underworld also called orcus and pluto&pluto\\ 
P69 (educated at)&10&where did jaren jackson senior play college basketball&georgetown university\\ 
P1877 (after a work by)&10&the movie catch me if you can is based on who&frank abagnale\\ 
\bottomrule
\end{tabular}
\caption{Relation with grounded QA pairs of Natural Questions Training Set (Top 1-82 by frequency).}
\label{tab:freq1}
\end{adjustwidth}
\end{table*}

\begin{table*}[ht]
\scriptsize
\centering
\begin{adjustwidth}{-10mm}{}
\begin{tabular}{l|c|l|l} \toprule
 \textbf{Relation}    & \textbf{Frequency} & \textbf{Question} & \textbf{Answer}\\ \midrule
P155 (follows)&10&what is the latest george rr martin book&a dance with dragons\\ 
P1029 (crew member)&10&who was the first to step on moon&neil armstrong\\ 
P3342 (significant person)&10&who was picked over kevin durant in the draft&greg oden\\ 
P749 (parent organization)&9&what does chi mean in chi st lukes&catholic health initiatives\\ 
P735 (given name)&9&who won in the war of alexander and porus&alexander\\ 
P463$_r$ (R: member of)&9&countries in the warsaw pact during the cold war&soviet union\\ 
P1376 (capital of)&8&cape town is the capital of what country&south africa\\ 
P156 (followed by)&8&the things we do for love song artist&10cc\\ 
P451 (unmarried partner)&8&who does elena date in the vampire diaries&stefan salvatore\\ 
P40 (child)&8&howard stark is the father of what superhero&iron man\\ 
P159 (headquarters location)&8&where is the head office of rbi located&mumbai\\ 
P287 (designed by)&8&who built the world first binary digit computer z1&konrad zuse\\ 
P551 (residence)&8&where did dorothy live in the wizard of oz&kansas\\ 
P647 (drafted by)&8&who does dwyane wade play for in the nba&miami heat\\ 
P30 (continent)&8&on what continents was the roman empire located at the height of its expansion&asia\\ 
P634 (captain)&7&who is the captain of kolkata knight riders&dinesh karthik\\ 
P828 (has cause)&7&what is the most common manifestation of portal hypertension -- induced splenomegaly&cirrhosis\\ 
P123 (publisher)&7&who made all the call of duty games&activision\\ 
P2408 (set in period)&7&when did hunchback of notre dame take place&1482\\ 
P27$_r$ (R: country of citizenship)&7&who was the last ruler of the tang dynasty&emperor ai of tang\\ 
P135$_r$ (R: movement)&7&who wanted the catholic church to reform and address&martin luther\\ 
P101$_r$ (R: field of work)&7&who invented the steam engine in the 1800s&james watt\\ 
P941 (inspired by)&6&who does squealer in animal farm represent in the russian revolution&vyacheslav molotov\\ 
P136$_r$ (R: genre)&6&who are the founding fathers of hip hop&grandmaster flash\\ 
P466$_r$ (R: occupant)&6&where did the patriots play before gillette stadium&foxboro stadium\\ 
P119$_r$ (R: place of burial)&6&who is buried in the great mausoleum at forest lawn glendale&michael jackson\\ 
P88 (commissioned by)&5&who built the castle in just one day&toyotomi hideyoshi\\ 
P110 (illustrator)&5&scary stories to tell in the dark artist&stephen gammell\\ 
P1366 (replaced by)&5&the old greek city-state of byzantium was rebuilt and became known as&constantinople\\ 
P169 (chief executive officer)&5&who become the ceo indian it company wipro in 2016&abidali neemuchwala\\ 
P3279 (statistical leader)&5&who is the captain of argentina national team fifa world cup 2018&lionel messi\\ 
P2388 (leader's office)&5&who does the us department of justice report to&united states attorney general\\ 
P53$_r$ (R: family)&5&who began the first dynasty of egyptian rulers&narmer\\ 
P2522$_r$ (R: victory)&5&who won season 2 of food network star&guy fieri\\ 
P823 (speaker)&5&who wrote we shall fight on the beaches&winston churchill\\ 
P748 (appointed by)&5&who can appoint comptroller and auditor general of india&president of india\\ 
P1363 (points/goal scored by)&5&who scored the winning goal for england in the 1966 world cup final&geoff hurst\\ 
P22 (father)&5&who was the king after david in the bible&solomon\\ 
P1027 (conferred by)&5&who presents national film award traditionally in india&directorate of film festivals\\ 
P750 (distributed by)&5&who own the rights to the black panther movie&walt disney studios motion pictures\\ 
P825 (dedicated to)&5&who was the song candle in the wind written about&marilyn monroe\\ 
P974 (tributary)&5&a tributary flowing into the mississippi from the east is the&ohio river\\ 
P8031 (perpetrator)&4&who was the guy who shoot in las vegas&stephen paddock\\ 
P885 (river source)&4&what is the starting point of the mississippi river&lake itasca\\ 
P631 (structural engineer)&4&who designed the first tunnel under the river thames&marc isambard brunel\\ 
P17$_r$ (R: country)&4&what are the countries of the united arab emirates&sharjah\\ 
P98 (editor)&4&who was an abolitionist who published and autobiography and anti-slavery newspaper&frederick douglass\\ 
P737 (influenced by)&4&qbasic is the extension of which programming language&quickbasic\\ 
P206$_r$ (R: on lake)&4&where does the river mekong start and end&mekong delta\\ 
P2789 (connects with)&4&a ship traveling through the panama canal could be crossing from the&atlantic ocean\\ 
P740 (location of formation)&4&where did the beatles started their career as a band&liverpool\\ 
P4743 (animal breed)&4&what kind of dog is bo and sunny&portuguese water dog\\ 
P466 (occupant)&4&who used to play in the alamo dome&utsa roadrunners\\ 
P2868$_r$ (R: subject has role)&4&who is the commander in chief of military&president of the united states\\ 
P5053 (fastest lap)&4&who won the 2018 chinese formula 1 grand prix&daniel ricciardo\\ 
P106$_r$ (R: occupation)&4&who is the griot that sings the epic&balla fasséké\\ 
P50$_r$ (R: author)&4&what is the title of langston hughes 's first book of poetry&the weary blues\\ 
P118 (league)&4&what conference is ohio state in for football&big ten conference\\ 
P2416$_r$ (R: sport discipline)&4&who has the world record for the long jump&galina chistyakova\\ 
P1552 (has quality)&4&which metal does the word ' ferrous ' refer to answer in words not symbols&iron\\ 
P8111 (unit)&4&unit of measure for area of a triangle&square metre\\ 
P179 (part of the series)&4&which games are in crash bandicoot n sane trilogy&crash bandicoot\\ 
P131$_r$ (R: is located in)&4&what is the name of capital of andhra pradesh&amaravati\\ 
P7047 (enemy of)&4&who sent doomsday to the end of time&superman\\ 
P725$_r$ (R: voice actor)&3&who does the voice of the cat in the hat&martin short\\ 
P61$_r$ (R: discoverer or inventor)&3&who discover the simple microscope first time and when&zacharias janssen\\ 
P6 (head of government)&3&who was the founder of the mauryan empire&chandragupta maurya\\ 
P264$_r$ (R: record label)&3&this artist was signed in 1952 by atlantic and brought a string of hits&ray charles\\ 
P462 (color)&3&what color was the white house when it was built&white\\ 
P533 (target)&3&who was killed in the ides of march&julius caesar\\ 
P972 (catalog)&3&who is on the top ten most wanted&alexis flores\\ 
P1344 (participant in)&3&india 's first olympic medal win as a free nation&1948 summer olympics\\ 
P106 (occupation)&3&what did pete best play in the beatles&drummer\\ 
P1366$_r$ (R: replaced by)&3&what is the old name for south africa&union of south africa\\ 
P171$_r$ (R: parent taxon)&3&what type of organism is made up of prokaryotic cells&archaea\\ 
P1411 (nominated for)&3&who won best director at the academy awards&guillermo del toro\\ 
P8345 (media franchise)&3&what is the first star wars movie in the series&star wars\\ 
P1433 (published in)&3&the story of seven ages by william shakespeare&as you like it\\ 
P20$_r$ (R: place of death)&3&who was the explorer that reached the cape of good hope at the southern tip of africa&bartolomeu dias\\ 
P87 (librettist)&3&who wrote the libretto for dido and aeneas&nahum tate\\ 
P3764 (pole position)&3&who won the abu dhabi grand prix 2017&valtteri bottas\\ 
P559 (terminus)&3&what is the southern end of the appalachian trail&springer mountain\\ 
\bottomrule
\end{tabular}
\caption{Relation with grounded QA pairs of Natural Questions Training Set (Top 83-164 by frequency).}
\end{adjustwidth}
\end{table*}

\begin{table*}[ht]
\begin{adjustwidth}{-15mm}{}
\centering
\scriptsize
\begin{tabular}{l|c|l|l} \toprule
 \textbf{Relation}    & \textbf{Frequency} & \textbf{Question} & \textbf{Answer}\\ \midrule
P366 (use)&3&what did the chinese use oracle bones for&pyromancy\\ 
P706$_r$ (R: takes place in)&3&what seven countries make up the subcontinent of south asia&sri lanka\\ 
P610 (highest point)&3&what is the highest point in the pyrenees mountains in france&aneto\\ 
P461 (opposite of)&3&the results of dehydration reactions can be reversed by&hydration reaction\\ 
P467 (legislated by)&3&which group is responsible for adopting the declaration of independence&second continental congress\\ 
P272 (production company)&3&what is the tv show riverdale based off of&archie comics\\ 
P140$_r$ (R: religion)&3&who is the leader of the baptist denomination&thomas helwys\\ 
P1419 (shape)&3&what is the shape of the earth 's orbit around the sun&ellipse\\ 
P942$_r$ (R: theme music)&3&clubs who sing you 'll never walk alone&liverpool f.c.\\ 
P376 (planet)&3&this planet is home to the great red spot&jupiter\\ 
P5202 (adapted by)&3&who wrote the lyrics for the song my way&paul anka\\ 
P171 (parent taxon)&3&trees of the betel nut genus of palms&areca\\ 
P509 (cause of death)&3&what was the cause of the tollund man 's death&hanging\\ 
P527$_r$ (R: has part)&3&the bronchi are considered to be part of the&respiratory system\\ 
P2849 (produced by)&3&where does red blood cell formation occur in adults&bone marrow\\ 
P460 (said to be the same as)&3&the word zion is an ancient biblical term that referred to what city&jerusalem\\ 
P1346$_r$ (R: winner)&3&when did the philadelphia eagles last win the super bowl&super bowl lii\\ 
P2341 (indigenous to)&3&dogri language is spoken in which state of india&himachal pradesh\\ 
P355 (subsidiary)&3&the main agency under the department of homeland security that is responsible for border security is&u.s. customs and border protection\\ 
P457 (foundational text)&3&where does one look to find the powers of a corporation&articles of incorporation\\ 
P108$_r$ (R: employer)&3&who is the current ceo of mcdonald 's corporation&steve easterbrook\\ 
P1923$_r$ (R: participating team)&3&last time houston astros have been to the world series&2017 world series\\ 
P8345$_r$ (R: media franchise)&2&what star wars movie came out before the last jedi&the empire strikes back\\ 
P112$_r$ (R: founded by)&2&real name of raj chandra in rani rashmoni&babughat\\ 
P113$_r$ (R: airline hub)&2&what airline has its hub in charlotte nc&american airlines\\ 
P156$_r$ (R: followed by)&2&what is the origin of the coptic language&egyptian language\\ 
P137 (operator)&2&who owns the white house in washington dc&national park service\\ 
P1552$_r$ (R: has quality)&2&what physical quantity is a measure of the amount of inertia and object has&mass\\ 
P2175 (disease treated)&2&topiramate ( topamax trokendi ) is used to treat which of the following diseases&epilepsy\\ 
P25 (mother)&2&who is carries mother on days of our lives&anna dimera\\ 
P170$_r$ (R: creator)&2&when was beverly cleary 's first book published&henry huggins\\ 
P641$_r$ (R: sport)&2&where do the rocks from curling come from&ailsa craig\\ 
P451$_r$ (R: unmarried partner)&2&who does raven end up with in the comics&beast boy\\ 
P4584 (first appearance)&2&what was the first game waluigi was in&mario tennis\\ 
P2670 (has parts of the class)&2&what do you rest a golf ball on&tee\\ 
P1040 (film editor)&2&who is the director of the film avatar&james cameron\\ 
P1056$_r$ (R: material produced)&2&who introduced the first micro processor in 1971&intel\\ 
P1192$_r$ (R: connecting service)&2&where does the eurostar leave from in paris&gare du nord\\ 
P1830 (owner of)&2&where do the carolina panthers play home games&bank of america stadium\\ 
P241$_r$ (R: military branch)&2&who served as the general of confederate forces during the civil war&robert e. lee\\ 
P111 (measure of)&2&joule is unit of . in mks system&energy\\ 
P19$_r$ (R: place of birth)&2&who was the last person to live in versaille&louis xvi\\ 
P291 (place of publication)&2&where was the institutes of the christian religion published&basel\\ 
P1056 (material produced)&2&by product of saponification of fats and oils&soap\\ 
P140 (religion)&2&of which religion is the avesta a sacred book&zoroastrianism\\ 
P137$_r$ (R: operator)&2&where do the fisher cats play in nh&northeast delta dental stadium\\ 
P162$_r$ (R: producer)&2&producer and director of silence of the lambs&edward saxon\\ 
P1582 (fruit of (taxon))&2&a plant that produces a type of bean&fabaceae\\ 
P286 (head coach)&2&2 ) who is the current manager of liverpool fc&jürgen klopp\\ 
P118$_r$ (R: league)&2&which nrl teams have never won a premiership&new zealand warriors\\ 
P413 (fielding position)&2&what position did ryan tannehill play in college&quarterback\\ 
P35 (head of state)&2&the longest serving samma ruler in sindh was&jam nizamuddin ii\\ 
P3173 (offers view on)&2&where is the leaning tower of pisa in italy located&pisa\\ 
P7959 (historic county)&2&archipelago that includes neolithic settlement of skara brae&orkney\\ 
P598$_r$ (R: commands)&2&union generals civil war army of the potomac&ambrose burnside\\ 
P306 (operating system)&2&what operating system does the macbook pro have&macos\\ 
P101 (field of work)&2&what did robert moog contribute to the music industry in the 1960s&electronic music\\ 
P27 (country of citizenship)&2&where is the actress that played wonder woman from&israel\\ 
P463 (member of)&2&what band is the girl from the grinch in&the pretty reckless\\ 
P4969 (derivative work)&2&what is the first book of pretty little liars&pretty little liars\\ 
P19 (place of birth)&2&where did anakin live before he met qui-gon&tatooine\\ 
P3938 (named by)&2&who developed the concept of an iron law of wages&ferdinand lassalle\\ 
P157$_r$ (R: killed by)&2&who does sansa end up with in game of thrones&ramsay bolton\\ 
P607 (conflict)&2&what battle did the tuskegee airmen help win&world war ii\\ 
P366$_r$ (R: use)&2&what kind of wax are crayons made from&paraffin wax\\ 
P551$_r$ (R: residence)&2&who lived in the land of nod east of eden&cain\\ 
P113 (airline hub)&2&where does porter airlines fly from in toronto&billy bishop toronto city airport\\ 
P927$_r$ (R: anatomical location)&2&where do the ilium the ischium and the pubis meet&acetabulum\\ 
P1000 (record held)&1&who holds the world record for 100 meters&usain bolt\\ 
P2541 (operating area)&1&what states does the i pass work in&illinois\\ 
P4647 (place of first performan)&1&where does medea go at the end of the play&athens\\ 
P483 (studio)&1&where was the dark side of the moon recorded&abbey road studios\\ 
P197$_r$ (R: adjacent station)&1&where does the rocky mountaineer leave from in vancouver&pacific central station\\ 
P36$_r$ (R: capital)&1&what country is in between poland and lithuania&kaliningrad oblast\\ 
P1589$_r$ (R: lowest point)&1&which state is bordered to the north by the artic ocean&alaska\\ 
P669 (located on street)&1&what area of paris is the eiffel tower&champ de mars\\ 
P1478 (has immediate cause)&1&the united states ' war on terror began in the wake of which of the following events&september 11 attacks\\ 
P1269$_r$ (R: facet of)&1&the enlightenment idea of separation of powers included which branches of government&legislature\\ 
P2679 (author of foreword)&1&who wrote the current edition of the catechism&pope john paul ii\\ 
P669$_r$ (R: located on street)&1&where did the beatles take the abbey road picture&abbey road studios\\ 
P837$_r$ (R: day in year)&1&what are three other names for makar sankranti&magh bihu\\ 
P3113 (does not have part)&1&which element in group 1 is not an alkaline metal&hydrogen\\ 
\bottomrule
\end{tabular}
\caption{Relation with grounded QA pairs of Natural Questions Training Set (Top 165-246 by frequency).}
\end{adjustwidth}
\end{table*}

\begin{table*}[ht]
\begin{adjustwidth}{-15mm}{}
\centering
\scriptsize
\begin{tabular}{l|c|l|l} \toprule
 \textbf{Relation}    & \textbf{Frequency} & \textbf{Question} & \textbf{Answer}\\ \midrule
P7047$_r$ (R: enemy of)&1&who took out the governor 's eye on walking dead&michonne\\ 
P59$_r$ (R: constellation)&1&brightest star in the constellation lyra dan word&vega\\ 
P3092 (film crew member)&1&who pioneered animated movies with his short feature steamboat willie in 1928&walt disney\\ 
P2348$_r$ (R: time period)&1&the main port of axum was the red sea city of&adulis\\ 
P736 (cover art by)&1&who wrote all quite on the western front&erich maria remarque\\ 
P469 (lakes on river)&1&where does the water from the nile come from&lake victoria\\ 
P205 (basin country)&1&in what country would you find the yellow river&china\\ 
P921$_r$ (R: main subject)&1&who began the systematic study of political science&american political science review\\ 
P4934 (calculated from)&1&a quantity 15 m / s to the north is a measure of&velocity\\ 
P1411$_r$ (R: nominated for)&1&who won the first oscar for best actress&janet gaynor\\ 
P4147 (conjugate acid)&1&give the name and formula for the acid derived from the following anion chlorite&chlorous acid\\ 
P276$_r$ (R: location)&1&the area between the tigris and euphrates rivers&mesopotamia\\ 
P413$_r$ (R: fielding position)&1&who has the most clean sheets in the world&iker casillas\\ 
P710$_r$ (R: participant)&1&in the second punic war between carthage and rome carthage formed an alliance with&massylii\\ 
P2563$_r$ (R: superpower)&1&who taught defence against the dark arts in book number 5&dolores umbridge\\ 
P2596 (culture)&1&which american civilization was located in a rain forest&maya civilization\\ 
P1071$_r$ (R: location of creation)&1&where does the young ones develop in humans&uterus\\ 
P1535$_r$ (R: used by)&1&what programming language is used in microsoft access&visual basic for applications\\ 
P400 (platform)&1&what consoles can you play star wars battlefront on&xbox one\\ 
P4913 (dialect of)&1&what type of arabic is spoken in palestine&south levantine arabic\\ 
P1066 (student of)&1&who is the minister during the regime of chandragupta&chanakya\\ 
P3342$_r$ (R: significant person)&1&who went before michael jordan in the draft&hakeem olajuwon\\ 
P86$_r$ (R: composer)&1&who wrote the power of love celine dion&candy derouge\\ 
P1427 (start point)&1&where did the tour de france start in 1954&amsterdam\\ 
P3373 (sibling)&1&who is the older brother mario or luigi&mario\\ 
P2512$_r$ (R: series spin-off)&1&which came first family guy or american dad&family guy\\ 
P2505$_r$ (R: carries)&1&where does the appalachian trail cross the hudson river&bear mountain bridge\\ 
P5009 (complies with)&1&what type of port is used by flash drives&usb mass storage device class\\ 
P2094 (competition class)&1&what weight class did muhammad ali fight in&heavyweight\\ 
P1889$_r$ (R: different from)&1&what name is given to fats that are liquid at room temperature&oil\\ 
P7937 (form of creative work)&1&wagner 's tristan und isolde is an example of&opera\\ 
P522 (type of orbit)&1&what 's the orbit of the international space station&low earth orbit\\ 
P1303 (instrument)&1&what kind of bass does john cooper play&bass guitar\\ 
P737$_r$ (R: influenced by)&1&who are the members of 3 6 mafia&juicy j\\ 
P263 (official residence)&1&where did zeus spend most of his time&mount olympus\\ 
P201 (lake outflow)&1&where does the water from lake okeechobee drain&caloosahatchee river\\ 
P178$_r$ (R: developer)&1&operating system developed in 1969 at at\&t 's bell laboratories&unix\\ 
P1312 (has facet polytope)&1&what is the opposite side of a right angle triangle&hypotenuse\\ 
P20 (place of death)&1&where did omri build his new political capital&samaria\\ 
P2936 (language used)&1&what is the national language of saudi arabia&arabic\\ 
P460$_r$ (R: said to be the same as)&1&what color is a school bus yellow or orange&chrome yellow\\ 
P682$_r$ (R: biological process)&1&which protein is responsible for the breakdown of a fibrin clot&plasmin\\ 
P3300 (musical conductor)&1&who did the music for ready player one&alan silvestri\\ 
P547 (commemorates)&1&name of ship that landed at plymouth rock&mayflower\\ 
P2079 (fabrication method)&1&the medium of the artwork that decorates the sistine chapel ceiling is&fresco\\ 
P1037 (director / manager)&1&who led the red shirts to victory in sicily&giuseppe garibaldi\\ 
P972$_r$ (R: catalog)&1&who is number one on america 's most wanted&jason derek brown\\ 
P263$_r$ (R: official residence)&1&which greek god ruled over a gloomy kingdom&hades\\ 
P2152 (antiparticle)&1&a packet or unit of light energy is called a&photon\\ 
P1462 (standards body)&1&who is responsible for creating the standards used on the internet&internet engineering task force\\ 
P664$_r$ (R: organizer)&1&when did they start using gloves in ufc&ufc 14\\ 
P937 (work location)&1&where did beethoven live most of his life&vienna\\ 
P4675$_r$ (R: appears in the form of)&1&what was robin 's name in batman and robin&dick grayson\\ 
P2596$_r$ (R: culture)&1&a ruined city on crete centre of the minoan bronze age civilisation&knossos\\ 
P2554 (production designer)&1&who made the movie all dogs go to heaven&don bluth\\ 
P1038 (relative)&1&what is the first name of huey 's dewey 's and louie 's uncle&donald duck\\ 
P3301 (broadcast by)&1&who is broadcasting the super bowl on sunday&nbc\\ 
P943 (programmer)&1&who wrote the first computer virus called elk cloner&rich skrenta\\ 
P30$_r$ (R: continent)&1&is puerto rico in north or central america&puerto rico\\ 
P135 (movement)&1&what kind of art did claude monet paint&impressionism\\ 
P5051 (towards)&1&which part of the cerebral hemisphere is supplied by the middle cerebral artery&cerebrum\\ 
P676$_r$ (R: lyrics by)&1&what beatles songs does paul play drums on&dear prudence\\ 
P364 (original language)&1&what language do they speak in kite runner&dari\\ 
P1071 (location of creation)&1&a town in the netherlands known for the production of a tin glazed earthenware&delft\\ 
P400$_r$ (R: platform)&1&name of the windows phone 8.1 virtual assistant&cortana\\ 
P452 (industry)&1&what did the hudson bay company do for canada&retail\\ 
P598 (commands)&1&who controlled or ordered the viet cong in combat&hoàng văn thái\\ 
P1303$_r$ (R: instrument)&1&who introduced the bass clarinet as a solo instrument in jazz&herbie mann\\ 
P3491 (muscle insertion)&1&what is the origin and insertion of the semimembranosus&medial condyle of tibia\\ 
P530 (diplomatic relation)&1&which two countries are on the western border of bolivia&chile\\ 
P1542 (has effect)&1&what disease is caused by bacterium treponema pallidum&syphilis\\ 
P1336 (territory claimed by)&1&the falkland islands are off the coast of what south american country&argentina\\ 
P747 (editions)&1&what is the latest ms office for mac&microsoft office 2016\\ 
P7153 (significant place)&1&on which island is the uss arizona memorial&honolulu\\ 
P610$_r$ (R: highest point)&1&the highest peak in north america mt . mckinley ( or denali ) is located in the state of&alaska\\ 
P1809 (choreographer)&1&who danced the lead role in appalachian spring&martha graham\\ 
P81 (connecting line)&1&what line is parsons green on tube map&district line\\ 
P122 (type of government)&1&what type of government did european settlers create in south africa in 1909&constitutional monarchy\\ 
P97$_r$ (R: noble title)&1&who was crowned the first holy roman emperor&charlemagne\\ 
P4552 (mountain range)&1&what mountain range is the blue mountains part of&great dividing range\\ 
P658 (tracklist)&1&what was u2 's lead single from ' the joshua tree '&with or without you\\ 
P195 (collection)&1&where is the original star spangled banner located&national museum of american history\\ 
P609 (terminus location)&1&where does route 66 start on the east coast&chicago\\ 
\bottomrule
\end{tabular}
\caption{Relation with grounded QA pairs of Natural Questions Training Set (Top 247-329 by frequency).}
\label{tab:freq2}
\end{adjustwidth}
\end{table*}

\begin{table*}[ht]
\begin{adjustwidth}{-15mm}{}
\centering
\scriptsize
\begin{tabular}{l|l} \toprule
Triplet & <'edward heath', ?,  "admiral's cup"> \\
Question & <mask> of edward heath which 1971 the british prime minister, edward heath, captained one of the winning boats. recent history.\\
True Passage & ...he captained britain's winning team for the \textbf{admiral's cup} in 1971 – while prime minister – and also captained the team in the 1979 fastnet race... \\ 
Pred. Relation & participant in: 0.33, winner: 0.19, participant: 0.04, victory: 0.03, sport: 0.03 \\
Pred. Answer &  admiral's cup (\cmark) \\
\midrule

Triplet & <'scary stories to tell in the dark', 'P110 (illustrated  by)', 'stephen gammell'> \\
Question & <mask> of scary stories to tell in the dark which evocative, nightmarish illustrations for alvin schwartz's "scary stories to tell in the dark" trilogy, he has illustrated nearly seventy \\
True Passage & scary stories to tell in the dark is a series of three collections of short horror stories for children, written by alvin schwartz and originally illustrated by \textbf{stephen gammell}... \\ 
Pred. Relation & illustrator: 0.13, creator: 0.11, author: 0.07, editor: 0.02, notable work: 0.02 \\
Pred. Answer &  stephen gammell (\cmark)\\
\midrule

Triplet & <'heeley', ?, 'sheffield tramway'> \\
Question & <mask> of heeley which first routes, to attercliffe and carbrook, brightside, heeley, nether edge and owlerton opened between 1873 \\
True Passage & ...\textbf{sheffield's old tramway} stretched from sheffield city centre to woodseats and heeley was at a time the terminus... \\ 
Pred. Relation & located in the administrative territorial entity: 0.21, located in the administrative territorial entity: 0.14, location: 0.07, shares border with: 0.04, terminus: 0.03 \\
Pred. Answer &  old tramway (\xmark) \\
\midrule

Triplet & <'pablo goncálvez', ?, 'patricia miller (tennis)'> \\
Question & <mask> of pablo goncálvez which luisa, was the first victim of uruguayan serial killer pablo goncálvez, who suffocated the 26-year old to \\
True Passage & ...the victim was 26 years old, had a degree in history and a practicing teacher, and was the sister of the well-known tennis player \textbf{patricia miller}... \\ 
Pred. Relation & sibling: 0.29, relative: 0.13, spouse: 0.04, relative: 0.04, place of burial: 0.02 \\
Pred. Answer &  patricia miller (\cmark)\\
\midrule

Triplet & <'chai prakan district', 'P131$_r$ (R: located in the administrative territorial entity)', 'chai prakan', > \\
Question & <mask> of chai prakan district which, is home to the district headquarters of<mask><mask><mask><mask> district in the far north of<mask>iang m<mask> province \\
True Passage & ...\textbf{chai prakan} is divided into four sub-districts ("tambons"), which are further subdivided into 44 administrative villages ("muban")... \\ 
Pred. Relation & located in the administrative territorial entity: 0.43, capital: 0.42, contains administrative territorial entity: 0.04, different from: 0.02, contains settlement: 0.01 \\
Pred. Answer &  chai prakarn (\cmark)\\
\midrule

Triplet & <'gothic western', ?, 'lorin morgan-richards'> \\
Question & <mask> of gothic western which lifestyle and his series "the goodbye family" has been categorized as gothic western. in addition to his work, rich \\
True Passage & ...in the young adult series, "the goodbye family" by \textbf{lorin morgan-richards} has been considered gothic western with an element of humor... \\ 
Pred. Relation & genre: 0.92, movement: 0.02, field of work: 0.0, genre: 0.0, occupation: 0.0 \\
Pred. Answer &  lorin morgan-richards (\cmark)\\
\midrule

Triplet & <'quentin bell', 'P40 (child)', 'virginia nicholson'> \\
Question & has kid of quentin bell which her father was the writer and art historian quentin bell, nephew of \\
True Passage & ...they had three children: julian bell, an artist and muralist; cressida bell, a notable textile designer; and \textbf{virginia nicholson}, the writer of "charleston: a bloomsbury house... \\ 
Pred. Relation & child: 0.98, father: 0.0, student: 0.0, sibling: 0.0, relative: 0.0 \\
Pred. Answer &  virginia nicholson (\cmark)\\
\midrule

Triplet & <'take me back to london', 'P361 (part of)', 'no.6 collaborations project'> \\
Question & <mask> of take me back to london which the border" featuring cabello and cardi b, and "take me back to london" featuring stormzy \\
True Passage & ...it was released as the eighth single from sheeran's fourth studio album \textbf{"no.6 collaborations project" (2019)}... \\ 
Pred. Relation & part of: 0.88, performer: 0.07, lyrics by: 0.01, producer: 0.0, followed by: 0.0 \\
Pred. Answer &  "no.6 collaborations project (\cmark)\\
\midrule

Triplet & <'u.s. route 441 in georgia', ? , 'lakemont, georgia'> \\
Question & <mask> of u.s. route 441 in georgia which area between u.s. route 23/441 and<mask> rabun.<mask><mask> has a post office with zip code \\
True Passage & ...from there it passes through the blue ridge mountain communities of wiley, \textbf{lakemont}, and tiger, the latter of which includes... \\ 
Pred. Relation & terminus location: 0.15, terminus: 0.11, located in the administrative territorial entity: 0.08, terminus: 0.05, connects with: 0.03 \\ 
Pred. Answer &  wiley  (\xmark)\\
\midrule

Triplet & <'anjelica huston', 'P57$_r$ (R: directed by)', 'agnes browne'> \\
Question & <mask> of anjelica huston which irish romantic comedy-drama film directed, produced by, and starring anjelica huston, based on the book "the mammy" by brendan o \\
True Passage & ...her next directorial effort, the irish dramedy \textbf{"agnes browne" (1999)} —in which she also starred as the title character— was released to mixed reviews... \\ 
Pred. Relation & terminus location: 0.15, terminus: 0.11, located in the administrative territorial entity: 0.08, terminus: 0.05, connects with: 0.03 \\
Pred. Answer &  "agnes browne (\cmark)\\
\midrule

Triplet & <'cadillac eldorado', ?, 'oldsmobile toronado'> \\
Question & <mask> of cadillac eldorado which 1967, cadillac adopted its own version of the upp for the cadillac eldor<mask>, using the cadillac v8 engine. \\
True Passage & ...by 2000, the eldorado was the last of a dying breed: its buick riviera and \textbf{oldsmobile toronado} stablemates had been discontinued, as had its perennial rival the lincoln mark... \\ 
Pred. Relation & follows: 0.38, followed by: 0.05, brand: 0.04, based on: 0.02, subclass of: 0.02 \\
Pred. Answer &  oldsmobile toronado (\cmark) \\
\midrule

Triplet & <'corsican nuthatch', 'P138 (named after)', 'john whitehead (explorer)'> \\
Question & eponym of corsican nuthatch which82 where he discovered a bird new to science, the corsican nuthatch. white<mask> travelled in malacca, north borneo, \\
True Passage & ...the corsican nuthatch was discovered by the english collector \textbf{john whitehead} in june 1883 when he shot a specimen while on a trip in the corsican mountains... \\ 
Pred. Relation & named after: 0.97, discoverer or inventor: 0.01, named after: 0.0, different from: 0.0, place served by transport hub: 0.0 \\
Pred. Answer &  john whitehead (\cmark)\\
\midrule
Triplet & <'mutual information', ?, 'information content'> \\
Question & <mask> of mutual information which formula\_13 is also often used for the related quantity of mutual<mask>, many authors use a lowercase formula\_14 for \\
True Passage & ...it quantifies the "\textbf{amount of information}" (in units such as shannons (bits), nats or hartleys) obtained about one random variable through observing the other random variable... \\ 
Pred. Relation & subclass of: 0.48, different from: 0.08, opposite of: 0.06, subclass of: 0.05, said to be the same as: 0.03 \\
Pred. Answer &  information theory (\xmark)\\ \midrule

Triplet & <'oculus (film)', 'P272 (production company)', 'intrepid pictures'> \\
Question & <mask> of oculus (film) which". in may 2012 filmdistrict acquired the film rights to what would become "oculus". soon after, the film released on april 11 \\
True Passage & ...eventually, \textbf{intrepid pictures} expressed interest in producing the film "as long as you don't do it found footage".... \\ 
Pred. Relation & production company: 0.29, producer: 0.19, distributed by: 0.07, screenwriter: 0.04, director: 0.03 \\
Pred. Answer &  intrepid pictures (\cmark) \\ 
\bottomrule
\end{tabular}
\caption{Examples of generated Relational QA datapoints and the predicted relation and answer by DPR pre-trained via \method.}
\label{tab:example}
\end{adjustwidth}
\end{table*}

\begin{table*}[ht]
\centering
\scriptsize
\begin{adjustwidth}{-22mm}{}
\begin{tabular}{l|c|l|l|l|l} \toprule
 \textbf{Relation Name}    & \textbf{Freq} & \textbf{Question} & \textbf{True Answer} & \textbf{\method Prediction} & \textbf{Supervised DPR Prediction} \\ \midrule
R: based on&0&theme song to bridge on the river kwai&the river kwai march&the river kwai march& march\\ 
subject has role&0&phenothiazines such as chlorpromazine were the first type of& antipsychotic& psychiatry& medication\\ 
practiced by&0&who does the call to prayer in islam&muezzin&muezzin& mosque\\ 
industry&1&what product or market does netflix deal with& streaming media& streaming media&netflix\\ 
made from&2&mohair is made from the fleece of what animal& angora goat& angora goat& goat\\ 
offers view on&2&where is the leaning tower of pisa built& pisa& pisa& pisa the leaning tower of pisa\\ 
R: residence&2&who is the founder of ramoji film city& ramoji rao& ramoji rao& telugu film producer ramoji rao\\ 
mother&2&who bore abraham first son in the bible& hagar& sarah&yishma'el\\ 
R: employer&3&who is the youngest judge currently sitting on the u.s. supreme court& neil gorsuch& neil gorsuch& leonard i. garth\\ 
R: has part&3&corpora cavernosa and corpus spongiosum are anatomic structures of& penis& penis&corpus cavernosum\\ 
indigenous to&3&urdu is the official language of which state& pakistan& jharkhand& jammu and kashmir\\ 
river source&4&what is the source of the colorado river& la poudre pass& la poudre pass& colorado begins at la poudre pass\\ 
tributary&5&river that joins the severn near chepstow crossword& river wye& river lugg& lugg\\ 
R: family&5&who was the second ruler of the davidic monarchy& solomon& jeroboam& solomon's son, rehoboam\\ 
R: genre&6&who is considered by many to be the father of soul& james brown& james brown& sam cooke\\ 
R: genre&6&who brought surf music to a national audience& the beach boys& the beach boys& dean\\ 
parent organization&8&who owns flying j and pilot truck stops& pilot corporation& pilot corporation& berkshire hathaway\\ 
narrator&10&who plays the mom in cheaper by the dozen& bonnie hunt& bonnie hunt& kate\\ 
educated at&10&where did the gabbie show go to college& university of pittsburgh& university of pittsburgh& the university of pittsburgh\\ 
director&15&who did the movie i can only imagine& erwin brothers& the erwin brothers& bart millard\\ 
executive producer&16&who stars in the movie the quiet place& john krasinski& john krasinski& emily blunt\\ 
R: player of&16&pitt players in the nfl hall of fame& mike ditka& ruben brown& tony dorsett\\ 
R: player of&16&who was the captain when india played its first-ever odi& ajit wadekar& srinivasaraghavan& s\\ 
shares border with&16&what state is directly west of north dakota& montana& montana& manitoba\\ 
depicts&19&who raised the american flag on iwo jima& michael strank& ira hayes& rene gagnon, ira hayes\\ 
depicts&19&faces of the presidents on mt. rushmore& abraham lincoln& thomas jefferson& theodore roosevelt\\ 
R: notable work&19&who won the 2015 great british baking show& nadiya hussain& joanne wheatley& edd kimber\\ 
presenter&21&who presented gardeners world from 2008 to 2010& toby buckland& joe swift& carol klein and joe swift\\ 
presenter&21&who are the new hosts of british bake off& noel fielding& noel fielding& sandi toksvig\\ 
material used&22&what kind of meat is on a t-bone& beef& cut from the short loin& tenderloin\\ 
main subject&24&new york times co v sullivan held that there must be proof of&actual malice& malice& truth\\ 
instance of&29&what kind of money do they use in russia&kopeks& ruble or rouble& the russian ruble or rouble\\ 
instance of&29&how does a plane wing create lift which physics concept applies&force& newton's second law& reaction force\\ 
country of origin&37&who used the springfield rifle in the civil war& united states& marine corps& army\\ 
R: subclass of&44&waste water that contain solid and liquid excreta refers to& sewage& sewage& pathogens\\ 
R: subclass of&44&when a blood vessel is injured the first phase in hemostasis to occur is& coagulation& wound healing& endothelial injury\\ 
located in&46&what part of new york is coney island& brooklyn& brooklyn& borough of brooklyn\\ 
R: part of&49&what regions of south asia have the highest population densities& philippines& philippines& indonesia\\ 
R: part of&49&what led to the downfall of the incan empire& battle of cajamarca& captured& victory\\ 
characters&50&the settlement of the israelites in canaan is the theme of which book& joshua& the book of joshua&book of joshua\\ 
R: award received&51&most number of national awards for best actress& shabana azmi& five& three\\ 
participating team&52&who did melbourne beat in the 1964 grand final& collingwood football club& collingwood& melbourne football club\\ 
spouse&53&who does jackson end up with in sons of anarchy& tara knowles& tara knowles& opie winston\\ 
R: instance of&56&which of the following is the si unit for length&metre&meter& litre\\ 
R: instance of&56&what is the most abundant neurotransmitter in the nervous system& serotonin& serotonin& glutamate\\ 
R: instance of&56&pricing tactics lower the price of a product below cost& loss leader& loss leader& increase in profits\\
named after&64&who was saint patrick's day named after& saint patrick& saint patrick&saint patrick\\ 
part of&70&arabian sea is the part of which ocean& indian ocean& northern indian ocean& the northern indian ocean\\ 
part of&70&who produces the most tires in the world&lego&lego tire: lego tire a lego& lego blocks. lego\\ 
officeholder&87&what is the name of the governor of new jersey& phil murphy& phil murphy& democrat phil murphy\\ 
participant&88&what two groups were fighting in the chinese civil war& communist party of china& communist party of china& republic of china\\ 
participant&88&who played the superbowl halftime show last year& bruno mars& beyoncé& coldplay\\ 
participant&88&who came second in the overall ranked of the tour de france last year& rigoberto urán& rigoberto urán& chris froome\\ 
filming location&98&what city does the terminator take place in& los angeles& los angeles& hemdale\\ 
filming location&98&where was back to the future three filmed& monument valley& monument valley& jamestown, california\\ 
\bottomrule
\end{tabular}

\caption{Comparison of the prediction of DPR initialized by \method with DPR without pre-training. These are all samples that two models made different predictions, and the relation frequency in the training set is less than 100.}
\label{tab:compare}
\end{adjustwidth}
\end{table*}

\begin{table*}[ht]
\begin{adjustwidth}{-15mm}{}
\footnotesize
\centering
\begin{tabular}{l|c|c|c|c} \toprule
 \textbf{Question}    & \textbf{Predicted Answer} & \textbf{True Answer} & \textbf{Match?} & \textbf{Predicted Relation} \\ \midrule
Who played mr darling on andy griffith show & Denver Pyle & Denver Pyle & \cmark  & P175 (performer) \\ \midrule
Who voices flik in a bug's life & Dave Foley & Dave Foley & \cmark & P725 (voice actor)\\ \midrule
Who's the dad of blair waldorf's baby & Chuck & Chuck& \cmark & P26 (spouse) \\ \midrule
\multirow{2}{*}{Where do you think glaciers can be found today} & \multirow{2}{*}{rocky mountains} & mountain ranges & \multirow{2}{*}{\xmark} & \multirow{2}{*}{P31$_r$ (R: instance of)}  \\
& & on every continent & & \\ \midrule
When did ginny weasley join the quidditch team & half-blood prince & half-blood prince & \cmark& P674$_r$  (R: characters)  \\ \midrule
When does far cry 5 for ps4 come out & 2018 & march 27, 2018 & \xmark & P400 (game platform)\\ \midrule
\multirow{2}{*}{When do millennials end and gen z start} & mid-1990s to  & \multirow{2}{*}{mid-1990s} & \multirow{2}{*}{\xmark} & \multirow{2}{*}{P155 (preceded by)}\\ 
& mid-2000s & & & \\ \midrule
Who killed hotchner's wife in criminal minds & George Foyet & George Foyet & \cmark & P7047 (enemy of)  \\ \midrule
\multirow{2}{*}{Who said walk tall and carry a big stick} & u.s. president  & \multirow{2}{*}{theodore roosevelt} & \multirow{2}{*}{\xmark} & \multirow{2}{*}{P170 (creator)} \\
& theodore roosevelt & & & \\ \midrule
Who does the voice of sheen from jimmy neutron & Jeffrey Garcia & Jeffrey Garcia & \cmark& P734 (family name) \\  \midrule
How many seasons of gossip girl are there & 6 & 6  & \cmark & P527 (has part)\\ \midrule
What can be used to detect the charge of particles & ionization detectors & particle detector & \xmark & P279$_r$ (R: subclass of) \\ \midrule
Who was robin in the original batman series & Burt Ward & Burt Ward & \cmark & P161 (cast member) \\ \midrule
What is the song funky cold medina about & a love potion & a fictional aphrodisiac & \xmark & P138 (named after) \\ \midrule
What do you call a quarter pounder in france & royal cheese & royal cheese & \cmark  & P1889 (different from) \\ \midrule
Who developed the first alternating  & \multirow{2}{*}{Galileo Ferraris} & \multirow{2}{*}{Nikola Tesla} & \multirow{2}{*}{\xmark} & \multirow{2}{*}{P61 (inventor)}\\ 
current electric system & & & &  \\ \midrule
Who won s5 of rupaul's drag race & Jinkx Monsoon & Jinkx Monsoon & \cmark & P1346  (winner) \\ \midrule
When was the svalbard global seed vault built & 2006 & 2006 & \cmark & P88 (built for) \\ \midrule
\multirow{2}{*}{Who was the mother of dragons married to} & \multirow{2}{*}{Khal Drogo} & Dothraki Horselord  & \multirow{2}{*}{\xmark} & \multirow{2}{*}{P26 (spouse)} \\ 
& & Khal Drogo & &  \\ \midrule
Which organization sets monetary  &  \multirow{2}{*}{the federal reserve} & \multirow{2}{*}{the federal reserve}  & \multirow{2}{*}{\cmark} & \multirow{2}{*}{P1001 (jurisdiction)} \\ 
policy for the united states & & & & \\ \midrule
What season of the voice was miley cyrus on & eleventh & season 11 & \xmark & P179 (part of series) \\ \midrule
Upon which document in american history is the & united states declaration  & united states declaration & \multirow{2}{*}{\cmark} & \multirow{2}{*}{P144 (based on)} \\
language of the declaration of sentiments based & of independence & of independence & & \\ \midrule
What kind of car does dale earnhardt jr drive & chevrolet camaro & chevrolet & \xmark & P54 (played for) \\ \midrule
How many times did brazil win the fifa world cup & five & five & \cmark & P1344 (participant in) \\ \midrule
\multirow{2}{*}{Second life is an example of a} & massively multiplayer & \multirow{2}{*}{an online virtual world} & \multirow{2}{*}{\xmark} & \multirow{2}{*}{P31 (instance of)}\\
& online role-playing games & & & \\ \midrule
What percentage of the world's  & \multirow{2}{*}{22\%} & \multirow{2}{*}{22\%} &  \multirow{2}{*}{\cmark} & \multirow{2}{*}{P276$_r$ (R: located in)} \\ 
population lives in east asia & & & & \\ \midrule
From which body part shurpnakha drive her name & fingernails & fingernails & \cmark & P186 (ingredient) \\ \midrule
How many chapters does the gospel of john have & four & 21 &  \xmark & P527 (has part) \\ \midrule
Who sang the original always on my mind & b.j. thomas & gwen mccrae & \xmark & P175 (performer) \\ \midrule
Where does the amazon river start and finish & atlantic ocean & atlantic ocean & \cmark & P403 (watercourse outflow)  \\  \midrule
Who did dwayne wade play for last year & the miami heat & miami heat & \xmark & P647 (drafted by) \\ \midrule
Who owns the rights to the power rangers & Hasbro & Hasbro & \cmark & P127 (owned by) \\
\bottomrule
\end{tabular}
\caption{Predicted relations for those QA pairs in Natural Questions Valid Set that cannot be aligned to WikiData. }
\label{tab:rel}
\end{adjustwidth}
\end{table*}


\end{document}